\documentclass[journal]{IEEEtran}
\usepackage{amsmath,amsfonts}
\usepackage{algorithmic}
\usepackage{algorithm}
\usepackage{array}
\usepackage[caption=false,font=footnotesize,labelfont=sf,textfont=sf]{subfig}
\usepackage{textcomp}
\usepackage{stfloats}
\usepackage{url}
\usepackage{verbatim}
\usepackage{graphicx}
\usepackage{cite}
\usepackage{bm}
\usepackage{multirow}
\usepackage{hyperref}
\usepackage[table,xcdraw]{xcolor}
\usepackage[normalem]{ulem}
\useunder{\uline}{\ul}{}
\hypersetup{
	colorlinks=true,
	linkcolor=blue,
	filecolor=blue,      
	urlcolor=blue,
	citecolor=blue,
}
\DeclareMathAlphabet\mathbfcal{OMS}{cmsy}{b}{n}
\usepackage{orcidlink}

\begin{document}
	
	\title{Unsupervised Hyperspectral and Multispectral Image Blind Fusion Based on Deep Tucker Decomposition Network with \\ Spatial–Spectral Manifold Learning}
	
	\author{He Wang$^{\orcidlink{0009-0000-3221-7165}}$, Yang Xu$^{\orcidlink{0000-0003-3514-9705}}$,~\IEEEmembership{Member, IEEE}, Zebin Wu$^{\orcidlink{0000-0002-7162-0202}}$,~\IEEEmembership{Senior Member, IEEE}, and Zhihui Wei$^{\orcidlink{0000-0002-4841-6051}}$,~\IEEEmembership{Member, IEEE}
		\vspace{-0.2cm}
			\thanks{
				This work was supported in part by the National Natural Science Foundation of China under Grant U23B2006, Grant 62071233, and Grant 62471233, 
				in part by the Jiangsu Provincial Natural Science Foundation of China under Grant BK20211570, 
				in part by the Jiangsu Provincial Innovation Support Program under Grant BZ2023046,
				in part by the Jiangsu Provincial Key Research and Development Program under Grant BE2022065-2,
				in part by the Open Fundations of Jiangsu Province Engineering Research Center of Airborne Detecting and Intelligent Perceptive Technology under Grant JSECF2023-03.
				(\textit{Corresponding~author:~Yang~Xu.})
				
				He Wang, Zebin Wu, and Zhihui Wei are with the School of Computer Science and Engineering, Nanjing University of Science and Technology (NJUST), Nanjing 210094, China (e-mail: he\_wang@njust.edu.cn, zebin.wu@gmail.com, gswei@njust.edu.cn)
				
				Yang Xu is with the School of Computer Science and Engineering, Nanjing University of Science and Technology (NJUST), Nanjing 210094, China, also with the Geological Exploration Technology Institute of Jiangsu Province, Nanjing 210018, China, and also with the Jiangsu Province Engineering Research Center of Airborne Detecting and Intelligent Perceptive Technology, Nanjing 210049, China (xuyangth90@njust.edu.cn)
				
			}}
	
	\markboth{Journal of \LaTeX\ Class Files,~Vol.~, No.~, MM~YY}%
	{Shell \MakeLowercase{\textit{et al.}}: A Sample Article Using IEEEtran.cls for IEEE Journals}
	
	
	\maketitle

	\begin{abstract}
		Hyperspectral and multispectral image fusion aims to generate high spectral and spatial resolution hyperspectral images (HR-HSI) by fusing high-resolution multispectral images (HR-MSI) and low-resolution hyperspectral images (LR-HSI). However, existing fusion methods encounter challenges such as unknown degradation parameters, incomplete exploitation of the correlation between high-dimensional structures and deep image features. To overcome these issues, in this article, an unsupervised blind fusion method for hyperspectral and multispectral images based on Tucker decomposition and spatial spectral manifold learning (DTDNML) is proposed. We design a novel deep Tucker decomposition network that maps LR-HSI and HR-MSI into a consistent feature space, achieving reconstruction through decoders with shared parameter. To better exploit and fuse spatial-spectral features in the data, we design a core tensor fusion network that incorporates a spatial spectral attention mechanism for aligning and fusing features at different scales. Furthermore, to enhance the capacity in capturing global information, a Laplacian-based spatial-spectral manifold constraints is introduced in shared-decoders. Sufficient experiments have validated that this method enhances the accuracy and efficiency of hyperspectral and multispectral fusion on different remote sensing datasets. The source code is available at \url{https://github.com/Shawn-H-Wang/DTDNML}.
	\end{abstract}
	
	
	\vspace{-0.5cm}
	\section{Introduction}
	\IEEEPARstart{H}{yperspectral} image (HSI) is a special image with a large amount of continuous spectral band information. In recent years, HSI is widely used in computer vision and remote sensing due to its detailed description of spectral features, like terrain classification\cite{6698324},\cite{9693311,9237144}, target detection\cite{9178496,8668709}, change detection\cite{8648482,9469924} and so on. Compared with HSIs, multispectral images (MSIs) have limited spectral information. However, limited by the intrinsic optical system, most of the captured images are lack of spatial or spectral information. To enhance the spatial resolution while preserving spectral dimension of the low resolution HSI (LR-HSI), a typical and effective way is to fuse it with high spatial resolution image. In the last decades, numerous effective fusion methods have been proposed, which can be mainly divided into two categories: model-driven methods and deep learning-based methods\cite{LI2022102926}.
	
	Model-driven methods primarily analyze the inherent relationships within the input data and construct a coupled solvable mathematical model. These methods solve the model through optimization techniques to obtain the best solution, including pan-sharpening, matrix decomposition, tensor decomposition and so on\cite{DIAN202140},\cite{10035509}. Pan-sharpening aims to fuse a single-channel high-resolution panchromatic image (HR-PAN) with MSI, providing higher spatial resolution\cite{7284770}. However, this approach may lead to partial loss of spectral information and color distortion. Conversely, direct fusion of HSI and MSI incurs higher computational cost and faces challenges in extracting spatial-spectral features. Thus, the HSI-MSI coupled decomposition-reconstruction approach has gained attention, which mainly focuses on matrix decomposition and spectral unmixing to extract rich spectral information. However, unfolding a three-dimensional HSI into a two-dimensional representation often ignores the inherent high-order structural information of the data\cite{8358017, 8494792}. Thanks to the high-dimensional representation properties of tensors, researchers gradually focus on tensor-based HSI-MSI fusion methods in recent years. In order to effectively enhance the extraction of relevant features in high-dimensional data, they can be decomposed into subspace and processed with mapping relationships corresponding to different dimensions. Even though model-based fusion methods have achieved satisfied performance, there are pressing issues that need to be addressed, including weak generalization performance of the models, high computational cost, and difficulties in capturing complex features.
	
	Deep learning (DL) has gained significant attention from researchers in various fields due to its powerful learning capability and data generalization ability\cite{9449622},\cite{Wang_2022_CVPR},\cite{10101853}. These techniques employ deep networks to learn deep spectral and spatial features from LR-HSI and HR-MSI respectively, and fuse the learned features to generate HR-HSI. Compared to model-based approaches, this method exhibits superior performance. Supervised training often struggles in blind fusion tasks aimed at learning unknown degradation parameters. Current unsupervised fusion methods, while capable of incorporating network-based iterative processes, predominantly focus on extracting deep spectral prior from LR-HSIs and abundance information from HR-MSIs, neglecting a direct representation of the data. In addition, there is still a lack of research on effective compression of the spectral and spatial dimension based on tensor representation.
	
	Hence, in this article, we propose an unsupervised HSI-MSI blind fusion method based on deep Tucker decomposition and spatial-spectral manifold learning to effectively achieve dimensionality reduction of high-dimensional data. A tensor-based deep autoencoder is constructed by employing convolution layers into the iterative process of the tensor Tucker fusion model. An enhanced approach for HSI-MSI fusion is designed, leveraging deep Tucker decomposition and multi-scale feature fusion. We consider the core tensor as a learnable feature within a deep Tucker decomposition network, allowing the encoded features of LR-HSI and HR-MSI to be learned and fused at multiple scales. The LR-HSI and HR-MSI are reconstructed through the coupled decoder to learn the factor matrices corresponding to different modes. The HR-HSI is reconstructed using the learned core tensor and the shared mode factor matrices parameters of the decoders. Considering the limited capacity of convolutional networks to capture global information and the risk of overfitting with only a simple decoder during reconstruction, we introduce weighted manifold learning through constructing a spatial-spectral manifold structure of inputs to constrain low-dimensional information. The learnable parameters of network are updated through joint loss optimization. Our main contributions can be summarized as follows:
	
	\begin{enumerate}
		\item We propose a novel unsupervised deep network, which incorporates two coupled decoders with shared parameters, leveraging the degradation model and deep Tucker decomposition, for blind fusion of HSI and MSI.  
		\item We design a Core Tensor Fusion Network based on a U-shaped network to explore spatial-spectral features efficiently, and facilitate interactive fusion of spatial-spectral features at each layer through the spatial-spectral attention module.
		\item Inspired by the effectiveness of manifold learning, we introduce a weighted manifold learning regularization, which is constructed through generating Laplacian matrices from LR-HSI and HR-MSI respectively. It is used for preserving the detail spatial-spectral structure, and
		effectively ensures dimensionality reduction of the highdimensional data.
		\item Extensive experiments conducted on four datasets validate the superior performance of the proposed method compared to state-of-the-art approaches. We present a detailed analysis of the strengths and limitations of our proposed method.
	\end{enumerate}
	
	The rest of this article is organized as follows. Related works are detailed in Section II. In Section III and Section IV, we introduce our proposed method in detail. Section V presents the extensive experiments and analysis. Finally, Section VI summarizes this work.
	
	\vspace{-0.4cm}
	\section{Related Work}
	\IEEEpubidadjcol
	\subsection{Model Driven Based Methods}
	The rapid development of spectral unmixing methods prompted researchers to think about how to more reasonably decompose HSI into purer endmember and abundance information. A fast sharpening method \cite{6731532} was proposed based on unmixing, which utilizes unconstrained least squares algorithm to solve the endmember matrix and abundance matrix. Its innovation lies in applying the unmixing process to sub images rather than the entire data. Yokoya \textit{et al.} \cite{5982386} proposed the HSI super-resolution method of Coupled Nonnegative Matrix Factorization (CNMF). It used nonnegative matrix factorization to alternately decompose LR-HSI and HR-MSI images to achieve the extraction of endmember and abundance information respectively. Due to CNMF's application of NMF unmixing cycle replacement to LR-HSI and HR-MSI, it usually has a significant time cost. Yi \textit{et al.}\cite{8358017} exploited spatial-spectral correlation between MSI and HSI through the sparse coefficients of the HS patch matrix with low-rank property. Subsequently, Dong \textit{et al.}\cite{7438864} proposed a non negative structured sparse representation super-resolution algorithm for HSIs for joint estimation of spectral dictionaries and sparse encoding. Jin \textit{et al.} \cite{9565143} designed an unsupervised adversarial autoencoder unmixing network. They analysed the correlation of local pixels and modeled abundance matrix prior. The spatial information will be set into the adversarial training network to improve the robustness.
	
	Overall, obtaining pure endmember and abundance information through matrix decomposition is a relatively effective fusion scheme. However, hyperspectral image data is usually represented in three-dimensional form, and this type of method requires unfolding the data into a two-dimensional matrix, thereby damaging the inherent high-dimensional structural information of the data. Benefited by the superiority of tensor decomposition models, hyperspectral fusion methods based on tensor decomposition models have become more famous in recent years. Dian \textit{et al.}\cite{NLSTF} first proposed the NLSTF method, which is based on the Tucker decomposition model and utilizes non local spatial similarity to learn dictionary information corresponding to different groups of patterns. In order to explore the role of sparse tensor constraints in fusion performance, Li \textit{et al.}\cite{8359412} proposed a Coupled Sparse Tensor Factorization (CSTF) model based on Tucker decomposition, which utilizes the reconstruction of high spatial spectral correlation in HR-HSI to promote the sparsity of core tensors and achieves good fusion performance. In addition, Xu \textit{et al.}\cite{8948303} introduced higher-order tensor representation into fusion problem. They transferred a 3-D HSI into a 4-D tensor, and reconstructed the HR-HSI based on the relationships between the decomposed core tensors. Although the model based hyperspectral fusion super-resolution method has achieved promising results, the weak generalization performance, high computational cost, and difficulty in characterizing complex features of the model are urgent issues that need to be addressed.
	
	\vspace{-0.2cm}
	\subsection{Deep Learning Based Methods}
	Deep learning technology has strong learning performance and data generalization ability, and has attracted the attention of scholars in various fields in recent years. In hyperspectral fusion super-resolution tasks, CNN, Transformer and Residual networks\cite{9796466,ma2021learning},\cite{SSSRCNN} have been proved that they could learn deep spectral and spatial features from LR-HSI and HR-MSI respectively. The early deep fusion theory\cite{9153037,LIU20201} mainly adopted the idea of pan-sharpening, which sampled LR-HSI to the same scale as HR-MSI before further learning. However, only upsampling may cause a mismatch between spectral information and pixels, resulting in limited effectiveness. Compared to directly upsampling a single branch, the multi branch method adopts an alternative strategy to alleviate this problem, which involves gradually upsampling through deconvolution or pixel shuffling, where high-resolution information is injected into the corresponding scale. Yang \textit{et al.}\cite{rs10050800} proposed a dual branch convolutional neural network for extracting features from HSI and MSI, respectively, to address this issue. Unlike ordinary RGB images, HSIs have strong spectral characteristics, while CNN is mainly used to extract local spatial features in the image. In addition, Dian \textit{et al.}\cite{CNN-Fus} proposed a deep fusion method based on subspace and matrix decomposition, which solves the Sylvester equation by introducing a well trained CNN denoiser with better generalization performance to achieve high-performance fusion super-resolution tasks. 
	
	However, such supervised deep networks often require a large amount of data for training, and the model needs to be retrained for different batches of datasets, resulting in low generalization ability of the network. In order to solve this problem, Wang \textit{et al.}\cite{DBIN, EDBIN} used different scales of CNN to design different degenerate learning networks to build a unsupervised learning (DBIN) framework. Zhang \textit{et al.}\cite{9136736} introduced the idea of deep image prior based on unsupervised training framework and proposed a deep blind hyperspectral super-resolution model. This method proposed a learnable shared tensor code idea to achieve deep level feature mining of images. Meanwhile, for the fusion of MSI and PAN, Diao \textit{et al.}\cite{ZeRGAN}  introduced the GAN into a zero sample multi-scale training method.
	
	\vspace{-0.2cm}
	\subsection{Model Guided in DL Methods}
	Although deep networks have achieved good results in hyperspectral fusion super-resolution tasks, there is still a difficult problem of poor interpretability in deep learning. Under the premise of ensuring performance, in order to improve the interpretability of the network, the current main solutions are introduce mathematical models to guide the network. They are mainly divided into two aspects. On the one hand, it is necessary to combine optimization algorithms to construct an unfolded deep fusion network. Xie \textit{et al.}\cite{9559907} proposed a deep fusion network based on degenerate model expansion for the first time. This method uses the proximal gradient descent method and constructs an expanded fusion network framework using matrix operations. Cao \textit{et al.}\cite{9559907} use the convolutional sparse coding to adopt the model-driven method and alternative algorithm to design an interpretable deep network structure for pansharpening (PanCSC-Net). Similar to it, on the basis of the matrix decomposition model, Shen \textit{et al.}\cite{ADMM-HFNet} proposed ADMM-HFNet based on the ADMM optimization framework, and Liu \textit{et al.}\cite{9681709} proposed a model inspired autoencoder (MIAE) that can perform unsupervised learning. 
	
	On the other hand, it combines traditional models to characterize learnable parameters as trainable deep networks. Qu \textit{et al.}\cite{uSDN} first proposed an unsupervised sparse Dirichlet network (uSDN) based on the idea of spectral unmixing. This method learns endpoint information that satisfies the Dirichlet distribution through deep networks and is applied to hyperspectral fusion tasks. Yao \textit{et al.}\cite{10.1007/978-3-030-58526-6_13} proposed a coupled hyperspectral fusion super-resolution network (CuCANet) based on the cross attention mechanism, which combines the idea of CNMF with deep networks and can learn unknown spatial spectral degradation information. On this basis, Zheng \textit{et al.}\cite{HyCoNet} proposed a hyperspectral fusion super-resolution network coupled with autoencoders (HyCoNet), which adopts the idea of spectral unmixing and achieves efficient fusion super-resolution by sharing end elements and abundance information learning module parameters. Recently, Yang \textit{et al.}\cite{10115230} migrated multilinear feature mapping into tucker decomposition, and proposed an unsupervised deep tensor network for HSI and MSI fusion tasks. However, there is still a lack of research on the deep fusion network for dimensionality reduction based on tensor decomposition.
	
	\vspace{-0.2cm}
	\section{Problem Formulation}
	Hereinafter, scalars, vectors, matrices, and tensors are denoted by $x$, $\bf{x}$, $\bf{X}$, and ${\bm{\mathcal X}}$, respectively.
	Tensor is an extension of a matrix that can have an arbitrary number of dimensions. It is an ideal representation for high-dimensional data. An order-d tensor can be represented as $\mathbfcal{T} \in \mathbb{R}^{I_1 \times I_2 \times \ldots \times I_d}$. Each dimension of the tensor is referred to as a mode. The elements of the tensor are denoted as $\mathbfcal{T}_{i_1, i_2, \ldots, i_d}$, where $i_1, i_2, \ldots, i_d$ represent the coordinates in different dimensions.
	
	The unfolded matrix is obtained by unfolding a tensor along the \textit{i}-th mode, while preserving the size of the \textit{i}-th mode and vectorizing the remaining modes. It is denoted as $\mathbf{T}_{(i)} = \texttt{unfold}(\mathbfcal{T}, i)$, where $\mathbf{T}_{(i)} \in \mathbb{R}^{I_i \times (I_1\times I_2\times\ldots \times I_{i-1}\times I_{i+1}\times \ldots \times I_d)}$. Mode multiplication is an operation between different modes of a tensor. It is commonly denoted as $\times_i$ to represent the product of a tensor and a matrix along the \textit{i}-th mode. Specifically, given a matrix $\mathbf{M} \in \mathbb{R}^{J_i \times I_i}$ representing the core tensor, the product of the tensor $\mathbfcal{T}$ and the matrix M along the \textit{i}-th mode is calculated as
	\begin{equation}\label{mode_product}
		\mathbfcal{T}\times_i\mathbf{M}=\mathbf{M}\mathbf{T}_{\left(i\right)}\in\mathbb{R}^{I_1\times I_2\times...\times I_{i-1}\times J_i\times I_{i+1}\times\ldots\times I_d}.
	\end{equation}
	
	Under the tensor representation framework, the fusion of HSI-MSI involves the reconstruction of the HR-HSI tensor $\mathbfcal{Z}\in \mathbb{R}^{W\times H\times S}$ by integrating the LR-HSI tensor $\mathbfcal{X}\in \mathbb{R}^{w\times h\times S}$ and the HR-MSI tensor $\mathbfcal{Y}\in \mathbb{R}^{W\times H\times s}$. Both $\mathbfcal{X}$ and $\mathbfcal{Y}$ are 3-D tensors, representing the degraded and ideal dimensions of width, height, and spectral bands in the images, respectively. It is worth noting that the dimensions of $\mathbfcal{X}$, $\mathbfcal{Y}$, and $\mathbfcal{Z}$ satisfy the relationship $w \ll W$, $h \ll H$, and $s \ll S$ due to the inherent limitations of the hyperspectral imaging system. According to the degradation model between the two observed images, $\mathbfcal{X}$ and $\mathbfcal{Y}$ can be regarded as the spatial and spectral degraded version of the unknown $\mathbfcal{Z}$ respectively, which is modeled as: 
	\begin{equation}\label{degradation1}
		\mathbfcal{X}=\mathbfcal{Z}\times_1 \mathbf{P_1} \times_2 \mathbf{P_2} + \mathbfcal{N}_\mathbf{h}\text{,}
	\end{equation}
	\begin{equation}\label{degradation2}
		\mathbfcal{Y}=\mathbfcal{Z}\times_3 \mathbf{P_3} + \mathbfcal{N}_\mathbf{m}\text{,}
	\end{equation}
	where $\mathbf{P_1} \in \mathbb{R}^{w\times W}$, $\mathbf{P_2}\in \mathbb{R}^{h\times H}$ and $\mathbf{P_3}\in \mathbb{R}^{s\times S}$ represent the degenerate matrices of tensorised HR-HSI on each mode. $\mathbfcal{N}_\mathbf{h}$ and $\mathbfcal{N}_\mathbf{m}$ mean the additive noise. Thus, the basic fusion model can be represented as a denoising model which is
	\begin{equation}\label{denoising}
		\min_{\mathbfcal{Z}}\frac{1}{2}\Vert \mathbfcal{X}-\mathbfcal{Z}\times_1 \mathbf{P_1} \times_2 \mathbf{P_2} \rVert_F ^{2} + \frac{1}{2}\Vert \mathbfcal{Y}-\mathbfcal{Z}\times_3 \mathbf{P_3} \rVert_F ^{2}\text{.}
	\end{equation}
	
	\begin{figure*}
		\centering
		\setlength{\abovecaptionskip}{-0.15cm}
		\includegraphics[width=0.9\linewidth]{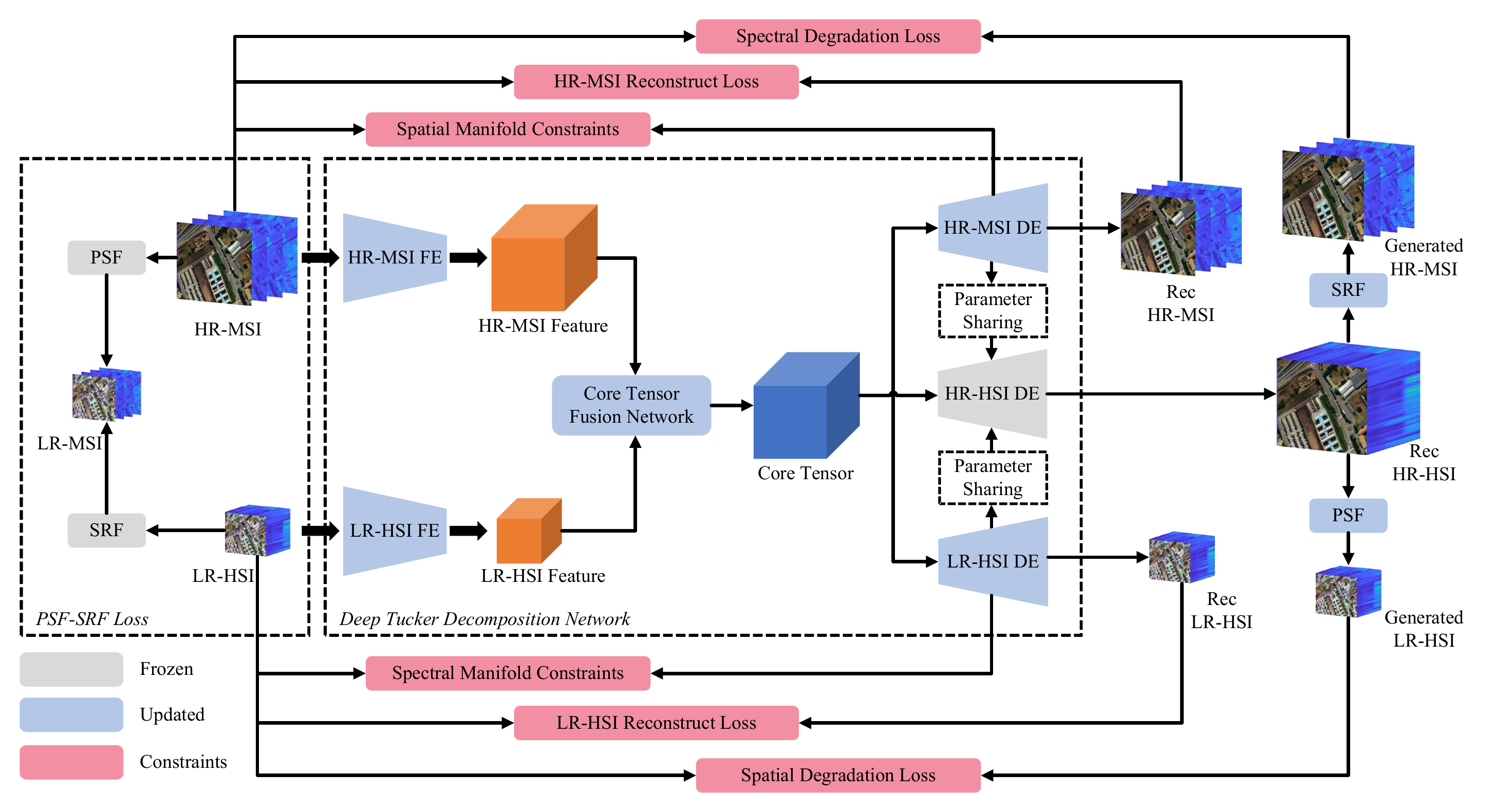}
		\caption{The architecture of the HSI-MSI blind fusion based on deep Tucker decomposition and manifold learning}
		\label{Architecture1}
	\end{figure*}
	
	However, the basic denoising model based on tensors often come with high computational costs. To address the issue of dimensional explosion, tensor decomposition is commonly employed. Tensor Tucker decomposition is a popular subspace decomposition method that decomposes a high-dimensional tensor into a core tensor and factor matrices corresponding to different modes. Through utilizing the decomposed components, essential structures and patterns in the data are captured to identify and select the most representative spectral information. Furthermore, the performance of the model \eqref{denoising} can be enhanced by utilizing Tucker decomposition\cite{8359412,8894531}. HR-HSI $\mathbfcal{Z}$ can be decomposed as
	\begin{equation}\label{HSI Tucker}
		\mathbfcal{Z}=\mathbfcal{C}\times_1\mathbf{W}\times_2\mathbf{H}\times_3\mathbf{S}\text{,}
	\end{equation}
	where $\mathbfcal{C}\in \mathbb{R}^{n_1\times n_2\times n_3}$ is the core tensor, $\mathbf{W}\in \mathbb{R}^{W\times n_1}$, $\mathbf{H}\in \mathbb{R}^{H\times n_2}$ and $\mathbf{S}\in \mathbb{R}^{S\times n_3}$ are the mode-1, mode-2 and mode-3 factor matrices respectively. Replace \eqref{degradation1} and \eqref{degradation2} with \eqref{HSI Tucker}, the LR-HSI $\mathbfcal{X}$ and HR-MSI $\mathbfcal{Y}$ can be represented as
	\begin{equation}\label{LR Tucker}
		\begin{aligned}
			\mathbfcal{X}&=\mathbfcal{C}\times_1(\mathbf{P_1}\mathbf{W})\times_2(\mathbf{P_2}\mathbf{H})\times_3\mathbf{S}\\
			&=\mathbfcal{C}\times_1\mathbf{W}^{*}\times_2\mathbf{H}^{*}\times_3\mathbf{S}\text{,}
		\end{aligned}
	\end{equation}
	
	\begin{equation}\label{MSI Tucker}
		\begin{aligned}
			\mathbfcal{Y}&=\mathbfcal{C}\times_1\mathbf{W}\times_2\mathbf{H}\times_3(\mathbf{P_3}\mathbf{S})\\
			&=\mathbfcal{C}\times_1\mathbf{W}\times_2\mathbf{H}\times_3\mathbf{S}^{*}\text{,}
		\end{aligned}
	\end{equation}
	where the core tensor $\mathbfcal{C}$ is shared.
	
	Thus, the HSI-MSI fusion model based on degraded model and Tucker decomposition can be transformed to calculate the optimal core tensor and mode-factor matrices, represented as
	\begin{equation}\label{base model}
		\mathop {\min }\limits_{\mathbfcal{C},\mathbf{W},\mathbf{H},\mathbf{S}}
		\begin{aligned}
			\frac{1}{2} \Vert \mathbfcal{X}-\mathbfcal{C}\times_1(\mathbf{P_1}\mathbf{W})\times_2(\mathbf{P_2}\mathbf{H})\times_3\mathbf{S} \rVert_F ^{2} +\\ \frac{1}{2} \Vert \mathbfcal{Y}-\mathbfcal{C}\times_1\mathbf{W}\times_2\mathbf{H}\times_3(\mathbf{P_3}\mathbf{S}) \rVert_F ^{2}\text{.}
		\end{aligned}
	\end{equation}
	To reconstruct the HR-HSI, we need to get optimal core tensor $\mathbfcal{C}$ and the corresponding factor matrices $\mathbf{W}$, $\mathbf{H}$, $\mathbf{S}$.
	
	\section{Methodology}
	According to \eqref{base model}, Alternating Least Squares (ALS) and Alternating Direction Method of Multipliers (ADMM) are typically employed for iterative optimization \cite{8941238,9737043}. However, shallow feature representations are insufficient for effectively extracting high-dimensional information, and phenomena such as overfitting may occur in sub-problem optimization. To overcome these issues, we treat the core tensor as a learnable and shareable feature extracted from the original data. Utilizing a multi-scale upsampling and downsampling network, our aim is to learn the mapping from the input image to the feature space, effectively capturing the spectral information from the LR-HSI and the spatial structural information from the HR-MSI. This approach significantly enhances the interpretability and fusion performance of the network. Additionally, through designing a reconstruction module with shared parameters, we can effectively learn the factor matrices associated with different modes in the tensor decomposition. Finally, weighted manifold constraints alone spatial and spectral modes are constructed for preserving detail spatial-spectral structure. Experimental results confirm the superiority of our proposed method over existing hyperspectral fusion super-resolution methods. The architecture of proposed method is shown in Fig.\ref{Architecture1}.
	
	\begin{figure*}
		\vspace{-0.2cm}
		\centering
		\setlength{\abovecaptionskip}{-0.15cm}
		\includegraphics[width=0.75\linewidth]{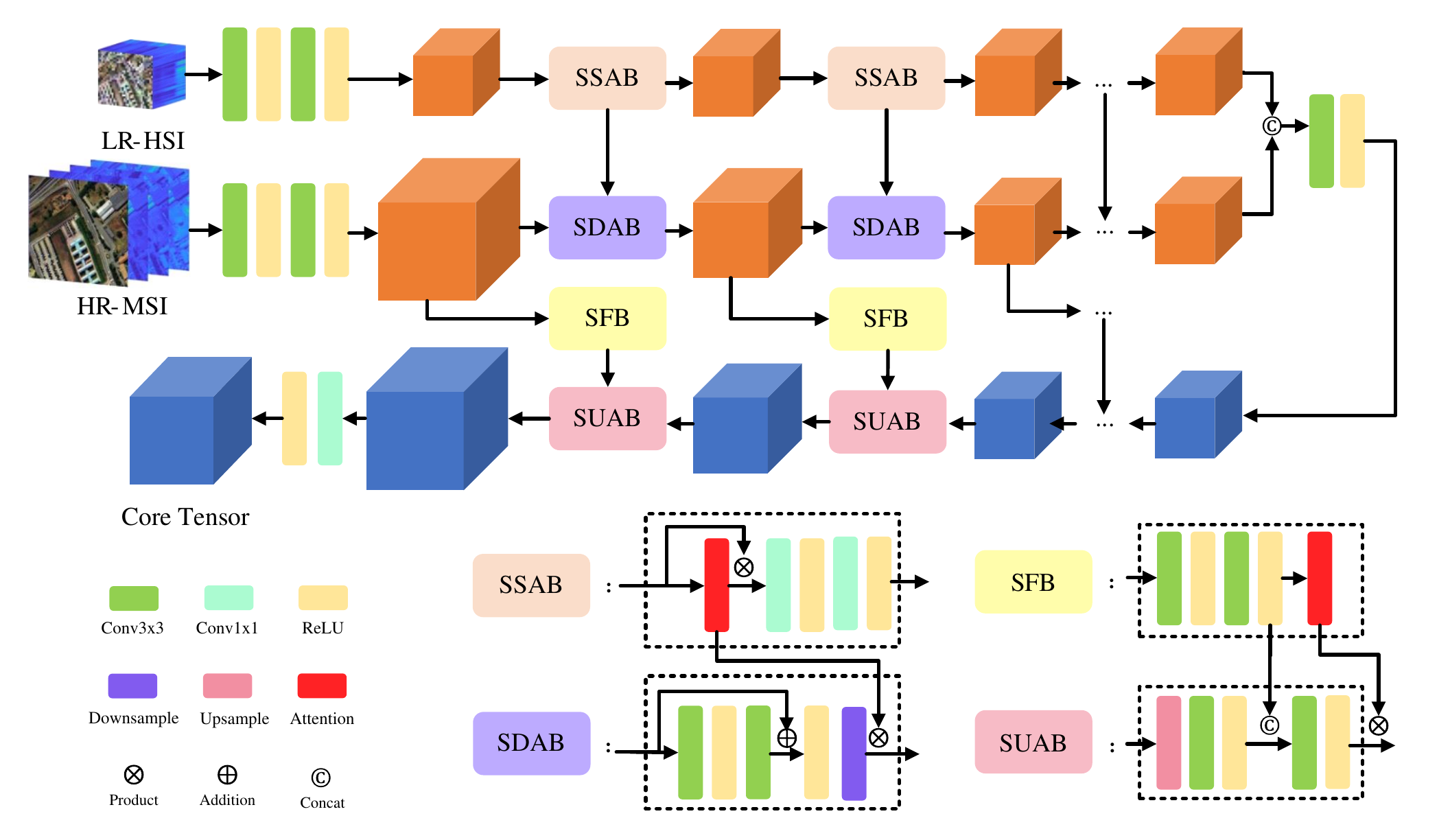}
		\caption{The structure of Core Tensor Fusion Network.}
		\label{DTDN}
	\end{figure*}
	
	\subsection{Deep Tucker Decomposition Network}
	From \eqref{LR Tucker} and \eqref{MSI Tucker}, it is shown that the core tensor, as shared information, contains spatial spectrum correlated information. 
	Therefore, we assume that the core tensor can be represented as a deep feature acquired from the input data, denoted as $F_\mathbfcal{C}$. This process is implemented through designing a fusion network, denoted as
	\begin{equation}\label{core_tensor}
		\mathbfcal{F}_{\mathbfcal{C}}=f_\text{fuse}(\mathbfcal{X},\mathbfcal{Y}; \Phi)\text{,}
	\end{equation}
	where $f_\text{fuse}$ and $\Phi$ represents a trainable fusion network and its parameters respectively.
	
	Due to the multilinear nature of tensor mode multiplication, we can learn the parameters of factor matrices by using 1x1 Conv layers, denoted as $f_{\text{rec}}$, based on the representation forms of factor matrices corresponding to different modes. By training the reconstruction network on LR-HSI and HR-MSI data, we can optimize the network parameters. Furthermore, considering the relationships among the factor matrices mentioned earlier, we share the weight parameters corresponding to different modes and achieve the reconstruction of LR-HSI, HR-MSI, and HR-HSI. The reconstructed images can be denoted as
	\begin{equation}\label{rec_lr_hsi}
		\hat{\mathbfcal{X}}=f_{\text{rec}}(\mathbfcal{F}_{\mathbfcal{C}}; \Theta_{\mathbf{S}}, \Theta_{\mathbf{w},\mathbf{h}})\text{,}
	\end{equation}
	\begin{equation}\label{rec_hr_msi}
		\hat{\mathbfcal{Y}}=f_{\text{rec}}(\mathbfcal{F}_{\mathbfcal{C}}; \Theta_{\mathbf{s}}, \Theta_{\mathbf{W},\mathbf{H}})\text{,}
	\end{equation}
	\begin{equation}\label{rec_hr_hsi}
		\hat{\mathbfcal{Z}}=f_{\text{rec}}(\mathbfcal{F}_{\mathbfcal{C}}; \Theta_{\mathbf{S}}, \Theta_{\mathbf{W},\mathbf{H}})\text{,}
	\end{equation}
	where $\hat{\mathbfcal{X}}$, $\hat{\mathbfcal{Y}}$ and $\hat{\mathbfcal{Z}}$ are the reconstructed LR-HSI, HR-MSI and HR-HSI respectively. The $\Theta$ represents learnable parameters of factor matrices. 

	\subsection{Core Tensor Fusion Network}
	One of the main challenges in the fusion of different modalities is how to learn the mapping of shared information between modalities. Although there are existing studies that have experimented with feature alignment, these methods have largely focused on natural images or medical images\cite{10145843}. In the context of HSI and MSI fusion, it is crucial to consider how to extract the corresponding spatial and spectral relationships\cite{9879770,10198668},\cite{9732243}. Thus, to achieve alignment and fusion of different modal features, we propose the Core Tensor Fusion Network (CTFN) based on attention and multi-scale feature fusion mechanisms 
	
	CTFN is a U-shaped fusion network. The network structure is illustrated as Fig.\ref{DTDN}. Firstly, the input image will be encoded through a feature encoding module, which contains two layers of convolutional neural networks (CNN) with a kernel size of $3\times3$, to obtain the relevant features for processing. We denote them followed by a ReLU activation layer as $f_i^{3\times3}\left(x\right)=ReLU\left(Conv\left(x\right)\right)$. The obtained features can be represented as
	\begin{equation}\label{Fx}
		\mathbfcal{F}_\mathbfcal{X}=f_{\text{LR-HSI-FE}}\left(\mathbfcal{X}\right)=f_2^{3\times3}\left(f_1^{3\times3}\left(\mathbfcal{X}\right)\right)\text{,}
	\end{equation}
	\begin{equation}\label{Fy}
		\mathbfcal{F}_\mathbfcal{Y}=f_{\text{HR-MSI-FE}}\left(\mathbfcal{Y}\right)=f_2^{3\times3}\left(f_1^{3\times3}\left(\mathbfcal{Y}\right)\right)\text{.}
	\end{equation}
	
	Although LR-HSI and HR-MSI contain sufficient spectral and spatial information, there is still a significant difference in the spatial dimensions between them. Therefore, we utilize downsample and upsample module to extrat multi-scale spatial information. To preserve the corresponding spectral information, we introduce the spectral squeeze attention block (SSAB), denoted as $f_{\text{SSAB}}$. It consists of $1\times1$ Conv layers and spectral attention mechanism, and allows for better extraction of spectral information from LR-HSI. The spatial downsampling block (SDAB), denoted as $f_{\text{SDAB}}$, consists of one basic ResBlocks and 2x2 Conv operations with a stride of 2. The number of blocks is determined by the spatial downsampling ratio of the input image, and defined as $s=\log_2(ratio)=\log_2(H/h)$. Meanwhile, the LR-HSI needs to go through an equal number of SSABs. After $s$ downsampling operations, the extracted features from HR-MSI and LR-HSI have the same size, we have
	\begin{equation}\label{f_ssab}
		\mathbfcal{F}_{\mathbfcal{X},s}=\underbrace{f_{\text{SSAB}}(f_{\text{SSAB}}(\dots f_{\text{SSAB}}}_{s}(\mathbfcal{F}_{\mathbfcal{X}})))\text{,}
	\end{equation}
	\begin{equation}\label{f_sdab}
		\mathbfcal{F}_{\mathbfcal{Y},s}=\underbrace{f_{\text{SDAB}}(f_{\text{SDAB}}(\dots f_{\text{SDAB}}}_{s}(\mathbfcal{F}_{\mathbfcal{Y}})))\text{.}
	\end{equation}
	These features are concatenated using the Concat operation, denoted as $[\mathbfcal{F}_{\mathbfcal{Y},s};\mathbfcal{F}_{\mathbfcal{X},s}]$. Then, a fusion block $f_{\text{fusion}}$, composed of a 3x3 Conv operation, is applied to perform the feature fusion. This operation can be represented as $\mathbfcal{F}_B = f_{\text{fusion}}([\mathbfcal{F}_{\mathbfcal{Y},s};\mathbfcal{F}_{\mathbfcal{X},s}])$. Subsequently, the fused feature is upsampled to generate the core tensor. The upsampling process is composed of a deconvolutional  upsampling block (SUAB), denoted as $f_{\text{SUAB}}$. The kernel size and stride of SUAB are both 2.
	
\begin{figure}[H]
		\centering
		\includegraphics[width=0.85\linewidth]{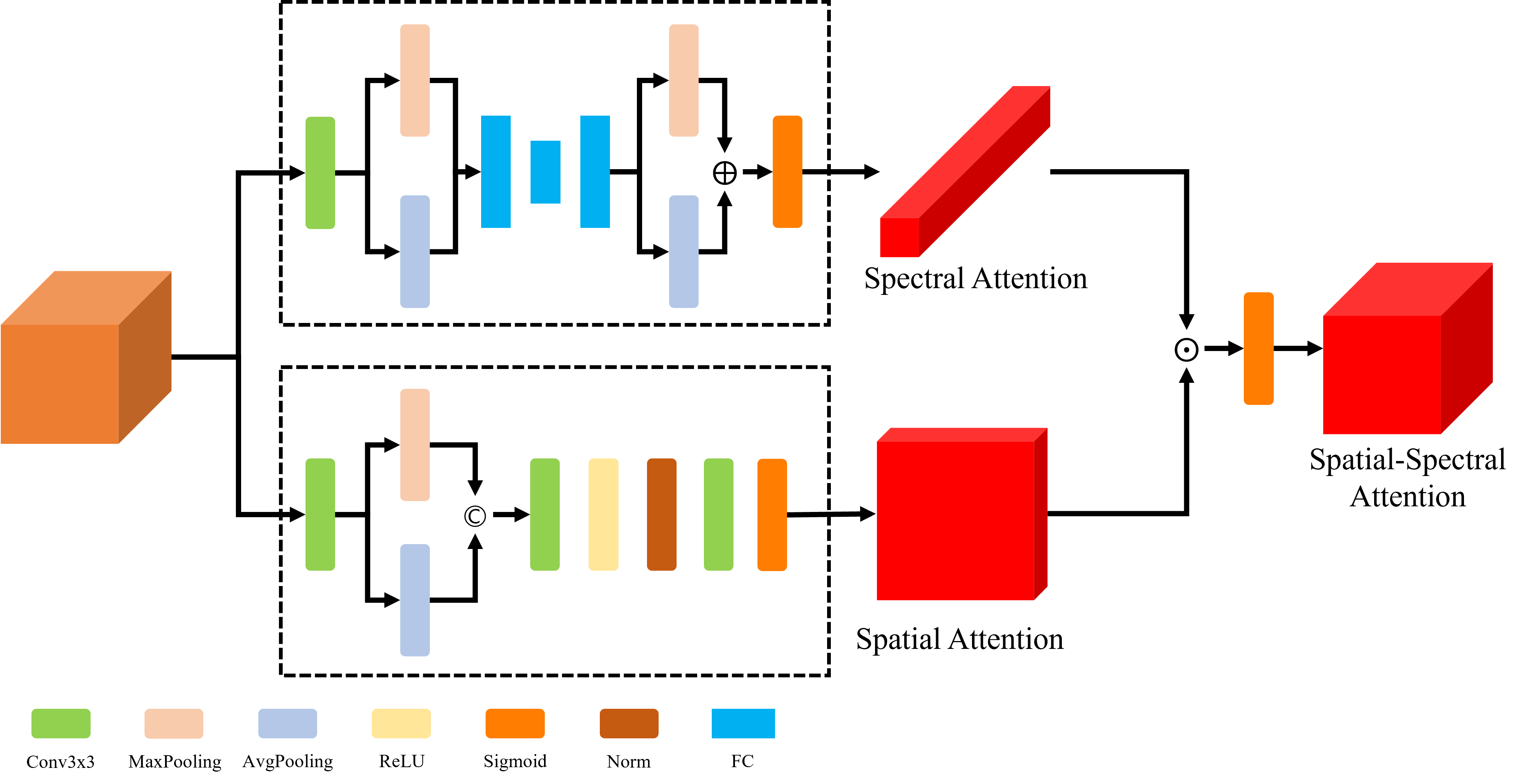}
		\caption{Spectral-Spatial Attention Module}
		\label{SSAM}
	\end{figure}

	During the multi-scale operating process, however, the upsampling and downsampling operations may cause the loss of some information in the features. To address this issue, the skip-connection fusion block (SFB) is constructed to facilitate information complementarity between features at corresponding scales and reduce information loss. From Fig.\ref{DTDN}, we first concatenate the features learned from $f_{\text{SFB}}$ and the upsampled feature $\mathbfcal{F}_{\text{up}}$. Inspired by channel and cross attention mechanisms in spectral feature extraction tasks\cite{LMFFNSSR, Mei2021CTFCNN}, we then incorporate a designed Spatial-Spectral Attention Module (SSAM), shown in Fig.\ref{SSAM}, into SSAB and SFB for the fusion of spatial spectral information. The $i$-th blocks interaction between SFB and SUAB can be represented as 
	\begin{equation}\label{Attention_employ}
		\begin{aligned}
			\mathbfcal{F}_{\text{up},i}=\mathbfcal{F}_{A,i} \otimes f^{3\times3}_2([f^{3\times3}_1(\mathbfcal{F}_{\text{up},i});f_{\text{SFB}}(\mathbfcal{F}_{\mathbfcal{Y},i})]) \text{,}
		\end{aligned}
	\end{equation}
	where $\mathbfcal{F}_{\mathbfcal{Y},i})$ is the downsampled feature with the same spatial size as $\mathbfcal{F}_{\text{up},i}$ and $\otimes$ means element-wise product.
	
	\vspace{-0.22cm}
	The SSAM incorporates both channel attention and spatial attention mechanisms to capture the preserved spectral and spatial information from the corresponding scale features obtained through downsampling. To reduce the loss of spectral information during compression, the channel attention mechanism is adopted. It preserves the spectral information from the LR-HSI. We denote the average pooling and maximum pooling as $AP$ and $MP$ respectively, the spectral attention map is defined as
	\begin{equation}\label{spe_A}
		\begin{aligned}
			\mathbf{F}_{A\text{-spe},i}=sigmoid(f_{fc}(AP(\mathbfcal{F}_{\mathbfcal{X},i}), MP(\mathbfcal{F}_{\mathbfcal{X},i})))\text{.}
		\end{aligned}
	\end{equation}
	It is applied to both the SSAB and SDAB by element-wise-multiplying it with the attention map to transmit the spectral information. In this process, the weights of the fully connected layers are shared, and we have $\mathbf{F}_{A\text{-spe},i}\in \mathbb{R}^{n_s\times1}$. The spatial attention is introduced to calculate the local-spatial weight from the high-resolution input data with convolution network. We define it as 
	\begin{equation}\label{spa_A}
		\begin{aligned}
			\mathbf{F}_{A\text{-spa},i}=sigmoid(f_{Norm}^{3\times3}[AP(\mathbfcal{F}_{\mathbfcal{Y},i});MP(\mathbfcal{F}_{\mathbfcal{Y},i})])\text{,}
		\end{aligned}
	\end{equation}
	where $\mathbf{F}_{A\text{-spa},i} \in \mathbb{R}^{\left(W/2i\right)\times\left(H/2i\right)}$, and $f_{Norm}^{3\times3}$ means basic Conv layer with normalization. According to \eqref{spe_A} and \eqref{spa_A}, we use broadcast element-wise product to construct the spatial-spectral attention map 
	\begin{equation}\label{spa_spe_A}
		\begin{aligned}
			\mathbfcal{F}_{A,i}=sigmoid(\mathbf{F}_{A\text{-}spa,i}\odot \mathbf{F}_{A\text{-}spe,i})\text{,}
		\end{aligned}
	\end{equation}
	where $\mathbfcal{F}_{A,i} \in \mathbb{R}^{\left(W/2i\right)\times\left(H/2i\right)\times n_s}$ and $i$ means the each stage of the U-like network. 
	
	The number of upsampling modules is equal to the number of downsampling modules, we have
	\begin{equation}\label{f_suab}
		\mathbfcal{F}_{\text{up}}=\underbrace{f_{\text{SUAB}}(f_{\text{SUAB}}(\dots f_{\text{SUAB}}}_{s}(\mathbfcal{F}_{\text{up},i})))\text{.}
	\end{equation}
	After upsampling, the obtained feature $\mathbfcal{F}_{\text{up}}$ needs to undergo a transform layer $f_{trans}$, achieving the spatial mapping of the core tensor, where $\mathbfcal{F}_{\mathbfcal{C}}=f_{trans}(\mathbfcal{F}_{up})$.
	
	\vspace{-0.2cm}
	\subsection{Weighted Manifold Learning}
	Despite the capability of convolutional neural networks to assist in learning deep features from training data, they still face challenges in capture of global detail information and dimensional reduction. To address these limitations and preserve the detail spatial-spectral correlations from HSIs, the Laplacian-manifold constraints is one of the efficient methods\cite{8948303},\cite{8253497},\cite{8469151}. It preserves both detail spatial and spectral structure. Commonly, the manifold structure is constructed by K-Nearest Neighbors (KNN) weighted Laplacian graph embedding. The weight of the adjacent elements is 
	\begin{equation}\label{spe_weight}
		\mathbf{A}(i,j)=\exp{(-\frac{\Vert {\bf{x}}_i-{\bf{x}}_j\Vert^2}{\sigma^2})}\text{,}
	\end{equation}
	where $i$ and $j$ means indices of correlated samples $\bf{x}$, and $\sigma$ determines the smoothness. Then, the Laplacian matrix $L$ is constructed as $\mathbf{L}=\mathbf{D}-\mathbf{A}$, where $\mathbf{D}=\texttt{diag}(\sum_{j=1}^{N}\mathbf{A}(i,j))$ represents a diagonal matrix with the diagonal elements being the sum of each row of the adjacency matrix. In Tucker decomposition, the factor matrices can be regarded as the low-dimensional space representation of the original tensor when unfolded along a certain mode. Assuming that the distances between samples in the factor matrices should also be minimized under this weight mapping. Inspired by this, we introduce spatial-spectral manifold constraint through constructing global Laplacian matrices. We expand and sample the input LR-HSI along the spectral dimension to construct a global adjacency matrix as $\mathbf{A}_S$, where the samples are $\mathbf{X}_{(3)}(i,:)$. According to \cite{6789755}, the spectral manifold constraint can be constructed with the learnable spectral-mode factor matrix, and denoted as $\mathcal{L}_{\text{spe-manifold}}=tr(\mathbf{S}^{\mathbf{T}} \mathbf{L}_S \mathbf{S})$, where $\mathbf{L}_S=\mathbf{D}_S-\mathbf{A}_S$.
	
	Similarly, the spatial weight matrix is constructed along spatial dimension based on HR-MSI. However, considering the pixel-wise unfolding, the desirable spatial weight matrix is large in size and computationally expensive, which is $\mathbf{A}_{\text{spatial}}\in\mathbb{R}^{HW\times HW}$. Moreover, it hinders the exploration of relationships among different factor matrices. To alleviate it, according to \cite{6671997}, we extend construction method of $\mathbf{A}_{S}$ to include two modes corresponding to the spatial dimensions. This approach can obtain samples related weights and extract spatial detail information from different modes. The spatial manifold constraints are constructed as $\mathcal{L}_{\text{spa-manifold}}=tr(\mathbf{W}^{\mathbf{T}}\mathbf{L}_W\mathbf{W})+tr(\mathbf{H}^{\mathbf{T}}\mathbf{L}_H\mathbf{H})$. Different with \cite{9094715}, instead of designing an optimization algorithm, it is automatically updated in conjunction with the joint loss to prevent overfitting of the decoding network during training. The manifold constraints are combined as shown in eq.\eqref{mc} 
	\begin{equation}\label{mc}
		\mathcal{L}_{\text{manifold}}={[\mathcal{L}_{\text{spe-manifold}};\mathcal{L}_{\text{spa-manifold}}]}.
	\end{equation}
	
	\subsection{Joint Loss Function}
	The loss function mainly consists of three main components: the reconstruction loss, the PSF-SRF loss and the manifold regularization.
	
	The Fig.\ref{Architecture1} shows how to reconstruct the LR-HSI and HR-MSI. In order to learn the core tensor and the factor matrices, it is usually trained by the basic denoising model like \eqref{denoising}. Thus, the reconstruction loss is defined as
	\begin{equation}\label{rec_loss}
		\begin{aligned}
			\mathcal{L}_{\text{rec}}=&\Vert\mathbfcal{X}-\hat{\mathbfcal{X}}\Vert_1 + \Vert\mathbfcal{Y}-\hat{\mathbfcal{Y}}\Vert_1\text{.}
		\end{aligned}
	\end{equation}
	
	Considered the unknown spatial and spectral degradation, namely point spread function (PSF) and spectral response function (SRF), an easy way to learn the parameters is designing a network. For the spatial degradation, we assume that the image of each band has the same spatial structure and can be seemed as suffering from the same spatial noise. Therefore, this process is simulated through a conv layer whose kernel size is $\texttt{scale}\times1$ and denoted as $f_{\text{PSF1}}(x;\Theta_\mathbf{P_1})$ and $f_{\text{PSF2}}(x;\Theta_\mathbf{P_2})$. For convenience of representation, we denote the PSF as $f_{\text{PSF}}(x;\Theta_\mathbf{P_1},\Theta_\mathbf{P_2})=f_{\text{PSF1}}(f_{\text{PSF2}}(x;\Theta_\mathbf{P_2});\Theta_\mathbf{P_1})$. Similarly, we adopt a similar approach to design the learning method for SRF parameters learning. Unlike PSF, SRF operates on the spectral dimension and follows an imaging principle that involves accumulating and normalizing the continuous spectral information along a certain position. This can be achieved by using a 1x1 Conv layer followed by a normalization layer. We denoted SRF as $f_{\text{SRF}}(x;\Theta_{\mathbf{P_3}})$. The basic degraded loss can be written as
	\begin{equation}\label{degraded1}
		\mathcal{L}_{\text{degraded}}=\Vert\mathbfcal{X}-f_{\text{PSF}}(\hat{\mathbfcal{Z}})\Vert_1+\Vert\mathbfcal{Y}-f_{\text{SRF}}(\hat{\mathbfcal{Z}})\Vert_1 \text{.}
	\end{equation}
	Some experiments have shown that basic degradation models often do not achieve satisfactory results, so it is necessary to introduce regularization constraints to improve the performance of the model. LR-MSI, a spatial degraded HR-MSI or a spectral degraded LR-HSI, has been proven to be a good intermediate medium for learning PSF and SRF. The discriminator of LR-MSI is designed as $\mathcal{L}_{\text{LR-MSI}}=\Vert f_{\text{SRF}}(\mathbfcal{X})-f_{\text{PSF}}(\mathbfcal{Y})\Vert_1$. The PSF-SRF loss is denoted as \eqref{PSF-SRF}:
	\begin{equation}\label{PSF-SRF}
		\mathcal{L}_{\text{PSF-SRF}}=\mathcal{L}_{\text{degraded}}+\gamma \mathcal{L}_{\text{LR-MSI}}\text{,}
	\end{equation}
	where $\gamma$ is the trade-off parameter of these to loss function.
	
	By combining all the aforementioned losses and regularized constraints, we jointly optimize the fusion network with the following loss function:
	\begin{equation}\label{joint-loss}
		\mathcal{L}=\mathcal{L}_{\text{rec}}+\alpha \mathcal{L}_{\text{PSF-SRF}}+\beta \mathcal{L}_{\text{manifold}}\text{,}
	\end{equation}
	where $\beta=[\beta_1, \beta_2]$ contains two super-parameters to constrain the spectral and spatial manifold respectively.
	
	\section{Experiment}
	
	\subsection{Datasets and Experiments Setup}
	We primarily utilized four hyperspectral remote sensing datasets for conducting simulated experiments and one dataset for real experiment. Pavia University dataset\cite{7946218} was acquired in the urban area of Pavia, Italy, using the Reflective Optics System Imaging Spectrometer (ROSIS) optical sensor. The image has dimensions of $610\times340\times103$, with a spatial resolution of 1.3 meters per pixel. The ROSIS sensor features 115 spectral bands, of which 103 bands remain after removing noisy bands. Chikusei dataset\cite{NYokoya2016} was captured in Chikusei, Japan, using the Headwall Hyperspec-VNIR-C sensor. It was made publicly available by Naoto Yokoya and Akira Iwasaki from the University of Tokyo. The dataset has dimensions of $2517\times2335\times128$, with a spatial resolution of 2.5 meters per pixel. Washington DC Mall (WaDC) hyperspectral image was captured through the Hyperspectral Digital Imagery Collection Experiment (HYDICE) sensor\cite{8738045}. It has dimensions of $1028\times307\times191$, with a spatial resolution of approximately 2.8 meters. The HYDICE sensor features 210 spectral bands, but after removing the bands affected by atmospheric opacity, the HSI contains 191 bands. SanDiego dataset\cite{xu2015anomaly} was captured by the Airborne Visible/Infrared Imaging Spectrometer (AVIRIS) sensor in the San Diego, USA. It has dimensions of $400\times 400 \times 224$, with a spatial resolution of approximately 20 meters. After removing water absorption bands, the number of the spectral band is 202. 
	
	All experiments are conducted to evaluate the performance of different fusion methods. First of all, we select the input data from them. For Pavia and WaDC, we crop an area with a size of $256\times 256$. Similarly, $512\times 512$ sized is selected from Chikusei. And for SanDiego, all the image is set as the reference HSI. During these experiments, we simulate spatial degradation and spectral degradation on the original HR-HSI. The original data served as a reference image for evaluating the fusion super-resolution effect. We generate HR-MSI for training process with known SRF. For the Pavia and SanDiego dataset, we use the Landsat-8 as the spectral degraded function. The SWIR-2 SRF was applied on the WaDC and Chikusei datasets. Additionally, we add Gaussian blur to the original data and performed sampling ratios (SR) of 4 and 8 to generate LR-HSI for training. To compare the experimental results among different methods, four different quality metrics were primarily employed for evaluation: Root Mean Square Error (RMSE), Peak Signal-to-Noise Ratio (PSNR), Spectral Angle Mapper (SAM)\cite{7946218}, Erreur Relative Globale Adimensionnelle de Synthèse (ERGAS)\cite{wald:hal-00365304}, Structural Similarity (SSIM)\cite{1284395} and Universal Image Quality Index (UIQI)\cite{995823}.
	
	The experiments are implemented using PyTorch, and conducted on an RTX 3090 24GB device. During the training process, the ADAM optimizer was employed with a learning rate of 5e-3, and the loss function was jointly optimized. The training was performed for 10000 epochs, and the learning rate was linearly decreased starting from the 3000th epoch. The network parameters were initialized using Kaiming initialization and Norm initialization. The hyperparameters $\alpha$, $\beta_1$, and $\beta_2$ were set to 1e-1, 1e-3, and 1e-2, respectively.
	
	\begin{figure*}
		\centering
		\setlength{\abovecaptionskip}{-0.15cm}
		\includegraphics[width=0.98\linewidth]{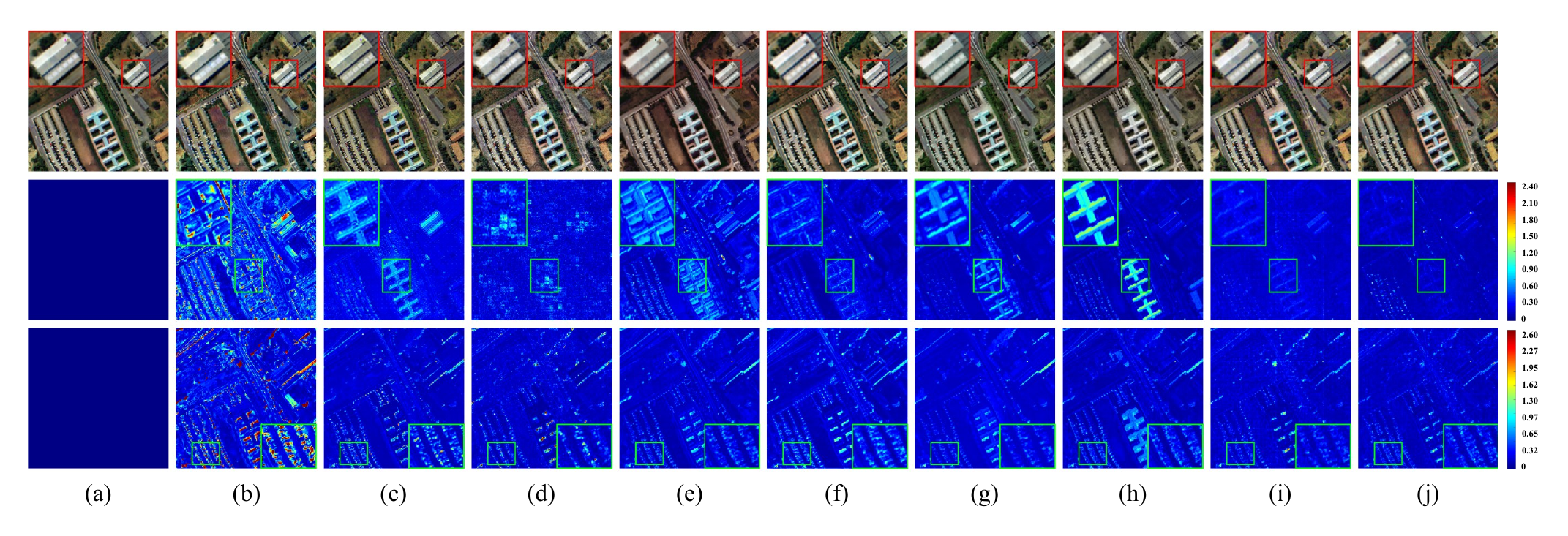}
		\caption{Illustration of fusion results on Pavia University dataset. First row: pseudo-RGB image (R:61, G:36, B:10). The second and third rows are heatmaps of RMSE and SAM between the ground truth and reconstructed respectively. The error map range are [0, 2.4] and [0, 2.6]. (a).GT, (b).CNMF\cite{5982386}, (c).HySure\cite{7000523}, (d).CSTF\cite{8359412}, (e).uSDN\cite{uSDN}, (f).CuCANet\cite{10.1007/978-3-030-58526-6_13}, (g).HyCoNet\cite{HyCoNet}, (h).MIAE\cite{9681709}, (i).UDTN\cite{10115230}, (j).DTDNML}
		\label{Pavia_compare}
	\end{figure*}
	
	\begin{figure*}
		\centering
		\setlength{\abovecaptionskip}{-0.15cm}
		\includegraphics[width=0.98\linewidth]{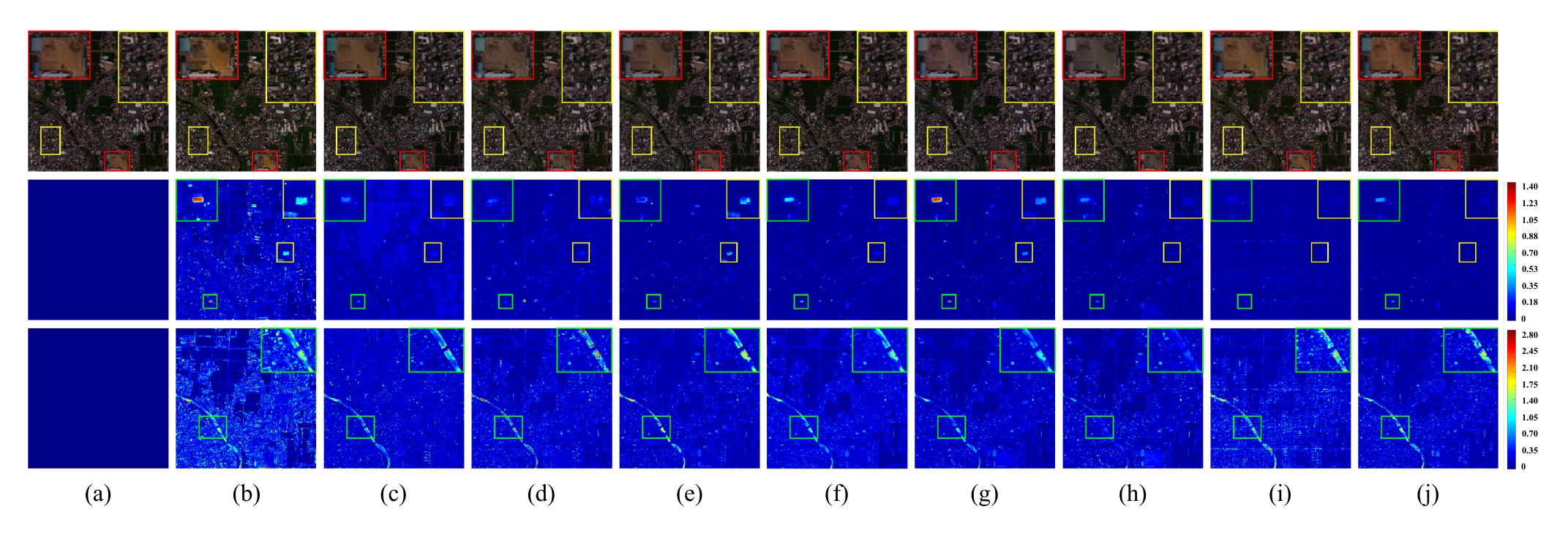}
		\caption{Illustration of fusion results on Chikusei dataset. First row: pseudo-RGB image (R:56, G:16, B:36). The second and third rows are heatmaps of RMSE and SAM between the ground truth and reconstructed respectively. The error map range are [0, 1.4] and [0, 2.8]. (a).GT, (b).CNMF\cite{5982386}, (c).HySure\cite{7000523}, (d).CSTF\cite{8359412}, (e).uSDN\cite{uSDN}, (f).CuCANet\cite{10.1007/978-3-030-58526-6_13}, (g).HyCoNet\cite{HyCoNet}, (h).MIAE\cite{9681709}, (i).UDTN\cite{10115230}, (j).DTDNML}
		\label{Chikusei_compare}
	\end{figure*}

	\subsection{Experimental Results}
	In this section, we compare our proposed method with other eight state-of-the-art hyperspectral fusion super-resolution methods: Coupled Nonnegative Matrix Factorization (CNMF)\cite{5982386}, Hyperspectral Subspace-based Regularized Fusion (HySure)\cite{7000523}, Coupled Sparse Tensor Factorization (CSTF)\cite{8359412}, Unsupervised Sparse Dirichlet-Net (uSDN)\cite{uSDN}, Coupled Unmixing Network with Cross-Attention (CuCANet)\cite{10.1007/978-3-030-58526-6_13}, Hyperspectral Coupled Convolution Network (HyCoNet)\cite{HyCoNet}, Model Inspired AutoEncoder (MIAE)\cite{9681709} and Unsupervised Deep Tensor Network (UDTN) \cite{10115230}. CNMF, HySure, and CSTF are model-based optimization methods implemented by MATLAB, while uSDN, CuCANet, HyCoNet, MIAE and UDTN are unsupervised training methods implemented by PyTorch. In this section, we bold and underline the optimal and suboptimal values respectively.
	
	\vspace{0.2cm}
	\textit{1) Pavia University}: In general, all the methods demonstrate promising fusion results. Specifically, within the model-driven methods, CSTF performs well by taking into account the high-dimensional structural correlation present in Pavia University dataset. However, traditional approaches rely on hand-crafted image priors, which leads to lower performance compared to deep learning-based methods. To facilitate a comprehensive comparison of fusion results, we present pseudo-color images, RMSE error maps, and SAM error maps generated by these fusion algorithms with sampling ratios of 8 as shown in Fig.\ref{Pavia_compare}. From the graphical representations, it is evident that tensor-based methods outperform the others in preserving both spatial and spectral information, with UDTN and MIAE being suboptimal. Moreover, we provide detailed quantitative analysis results of the nine methods in the Table \ref{tab:Tab1}. The empirical findings consistently support the superiority of DTDNML over other methods, as evidenced by its superior performance in key evaluation metrics such as PSNR and SAM. Notably, the most notable improvement is achieved at a sampling ratio of 4, where DTDNML achieves a PSNR of 36.41dB, representing an improvement of approximately 1dB. Similarly, at a sampling ratio of 8, the proposed method achieves 35.18dB of PSNR and reduces SAM to 4.23 degrees. However, it is important to acknowledge that in this case, DTDNML exhibits relatively weaker performance in overall SSIM when compared to other methods, primarily due to the challenge of fusing wavelength information in the presence of inconsistent illumination conditions. Additionally, HySure and MIAE demonstrate a strong ability to learn spectral features.
	\begin{table}[H]
		\caption{EVALUATION INDICES OF DIFFERENT HSI–MSI FUSION METHODS ON PAVIA UNIVERSITY DATASET}\label{tab:Tab1}
		\centering
		\resizebox{\linewidth}{!}{
			\begin{tabular}{ccccccc}
				\hline\hline
				\multicolumn{1}{c|}{Metrics}    & \multicolumn{1}{c|}{RMSE}            & \multicolumn{1}{c|}{PSNR}           & \multicolumn{1}{c|}{SAM}           & \multicolumn{1}{c|}{ERGAS}         & \multicolumn{1}{c|}{SSIM}           & UIQI           \\ \hline
				\multicolumn{1}{c|}{Optimal Value} & \multicolumn{1}{c|}{0}               & \multicolumn{1}{c|}{$+\infty$}      & \multicolumn{1}{c|}{0}             & \multicolumn{1}{c|}{0}             & \multicolumn{1}{c|}{1}              & 1              \\ \hline
				\multicolumn{7}{c}{SR=4}                                                                                                                                                                                                                      \\ \hline
				\multicolumn{1}{c|}{CNMF}       & \multicolumn{1}{c|}{0.0297}          & \multicolumn{1}{c|}{31.02}          & \multicolumn{1}{c|}{4.93}          & \multicolumn{1}{c|}{2.10}          & \multicolumn{1}{c|}{0.861}          & 0.983          \\
				\multicolumn{1}{c|}{HySure}     & \multicolumn{1}{c|}{0.0241}          & \multicolumn{1}{c|}{32.56}          & \multicolumn{1}{c|}{4.21}          & \multicolumn{1}{c|}{1.77}          & \multicolumn{1}{c|}{0.917}          & 0.987          \\
				\multicolumn{1}{c|}{CSTF}       & \multicolumn{1}{c|}{0.0180}          & \multicolumn{1}{c|}{35.27}          & \multicolumn{1}{c|}{4.27}          & \multicolumn{1}{c|}{{\ul 1.35}}    & \multicolumn{1}{c|}{0.944}          & 0.992          \\
				\multicolumn{1}{c|}{uSDN}       & \multicolumn{1}{c|}{0.0218}          & \multicolumn{1}{c|}{33.46}          & \multicolumn{1}{c|}{4.05}          & \multicolumn{1}{c|}{1.64}          & \multicolumn{1}{c|}{0.934}          & 0.988          \\
				\multicolumn{1}{c|}{CuCaNet}    & \multicolumn{1}{c|}{0.0201}          & \multicolumn{1}{c|}{34.36}          & \multicolumn{1}{c|}{3.91}          & \multicolumn{1}{c|}{1.53}          & \multicolumn{1}{c|}{0.949}          & 0.992          \\
				\multicolumn{1}{c|}{HyCoNet}    & \multicolumn{1}{c|}{0.0208}          & \multicolumn{1}{c|}{34.44}          & \multicolumn{1}{c|}{4.00}          & \multicolumn{1}{c|}{1.71}          & \multicolumn{1}{c|}{{\ul 0.951}}    & 0.992          \\
				\multicolumn{1}{c|}{MIAE}       & \multicolumn{1}{c|}{{\ul 0.0178}}    & \multicolumn{1}{c|}{{\ul 35.63}}    & \multicolumn{1}{c|}{{\ul 3.90}}    & \multicolumn{1}{c|}{1.42}          & \multicolumn{1}{c|}{0.948}          & {\ul 0.993}    \\
				\multicolumn{1}{c|}{UDTN}       & \multicolumn{1}{c|}{0.0183}          & \multicolumn{1}{c|}{34.93}          & \multicolumn{1}{c|}{4.85}          & \multicolumn{1}{c|}{1.47}          & \multicolumn{1}{c|}{0.930}          & 0.989          \\
				\multicolumn{1}{c|}{DTDNML}       & \multicolumn{1}{c|}{\textbf{0.0158}} & \multicolumn{1}{c|}{\textbf{36.41}} & \multicolumn{1}{c|}{\textbf{3.87}} & \multicolumn{1}{c|}{\textbf{1.24}} & \multicolumn{1}{c|}{\textbf{0.955}} & \textbf{0.994} \\ \hline
				\multicolumn{7}{c}{SR=8}                                                                                                                                                                                                                      \\ \hline
				\multicolumn{1}{c|}{CNMF}       & \multicolumn{1}{c|}{0.0672}          & \multicolumn{1}{c|}{24.04}          & \multicolumn{1}{c|}{9.56}          & \multicolumn{1}{c|}{4.64}          & \multicolumn{1}{c|}{0.879}          & 0.934          \\
				\multicolumn{1}{c|}{HySure}     & \multicolumn{1}{c|}{0.0317}          & \multicolumn{1}{c|}{30.11}          & \multicolumn{1}{c|}{4.83}          & \multicolumn{1}{c|}{2.41}          & \multicolumn{1}{c|}{\textbf{0.961}} & 0.949          \\
				\multicolumn{1}{c|}{CSTF}       & \multicolumn{1}{c|}{0.0266}          & \multicolumn{1}{c|}{31.63}          & \multicolumn{1}{c|}{4.91}          & \multicolumn{1}{c|}{2.01}          & \multicolumn{1}{c|}{0.882}          & 0.979          \\
				\multicolumn{1}{c|}{uSDN}       & \multicolumn{1}{c|}{0.0320}          & \multicolumn{1}{c|}{30.42}          & \multicolumn{1}{c|}{4.72}          & \multicolumn{1}{c|}{2.23}          & \multicolumn{1}{c|}{0.854}          & 0.972          \\
				\multicolumn{1}{c|}{CuCaNet}    & \multicolumn{1}{c|}{0.0227}          & \multicolumn{1}{c|}{33.19}          & \multicolumn{1}{c|}{5.08}          & \multicolumn{1}{c|}{1.76}          & \multicolumn{1}{c|}{0.938}          & 0.987          \\
				\multicolumn{1}{c|}{HyCoNet}    & \multicolumn{1}{c|}{0.0239}          & \multicolumn{1}{c|}{33.22}          & \multicolumn{1}{c|}{5.12}          & \multicolumn{1}{c|}{1.72}          & \multicolumn{1}{c|}{{\ul 0.949}}    & 0.987          \\
				\multicolumn{1}{c|}{MIAE}       & \multicolumn{1}{c|}{0.0245}          & \multicolumn{1}{c|}{34.25}          & \multicolumn{1}{c|}{4.70}          & \multicolumn{1}{c|}{2.06}          & \multicolumn{1}{c|}{0.943}          & 0.988          \\
				\multicolumn{1}{c|}{UDTN}       & \multicolumn{1}{c|}{{\ul 0.0202}}    & \multicolumn{1}{c|}{{\ul 34.47}}    & \multicolumn{1}{c|}{{\ul 4.50}}    & \multicolumn{1}{c|}{{\ul 1.52}}    & \multicolumn{1}{c|}{0.941}          & {\ul 0.989}    \\
				\multicolumn{1}{c|}{DTDNML}       & \multicolumn{1}{c|}{\textbf{0.0180}} & \multicolumn{1}{c|}{\textbf{35.18}} & \multicolumn{1}{c|}{\textbf{4.23}} & \multicolumn{1}{c|}{\textbf{1.40}} & \multicolumn{1}{c|}{0.938}          & \textbf{0.992} \\ \hline\hline
			\end{tabular}
		}
	\end{table}
	
	\begin{figure*}
		\centering
		\setlength{\abovecaptionskip}{-0.15cm}
		\includegraphics[width=0.98\linewidth]{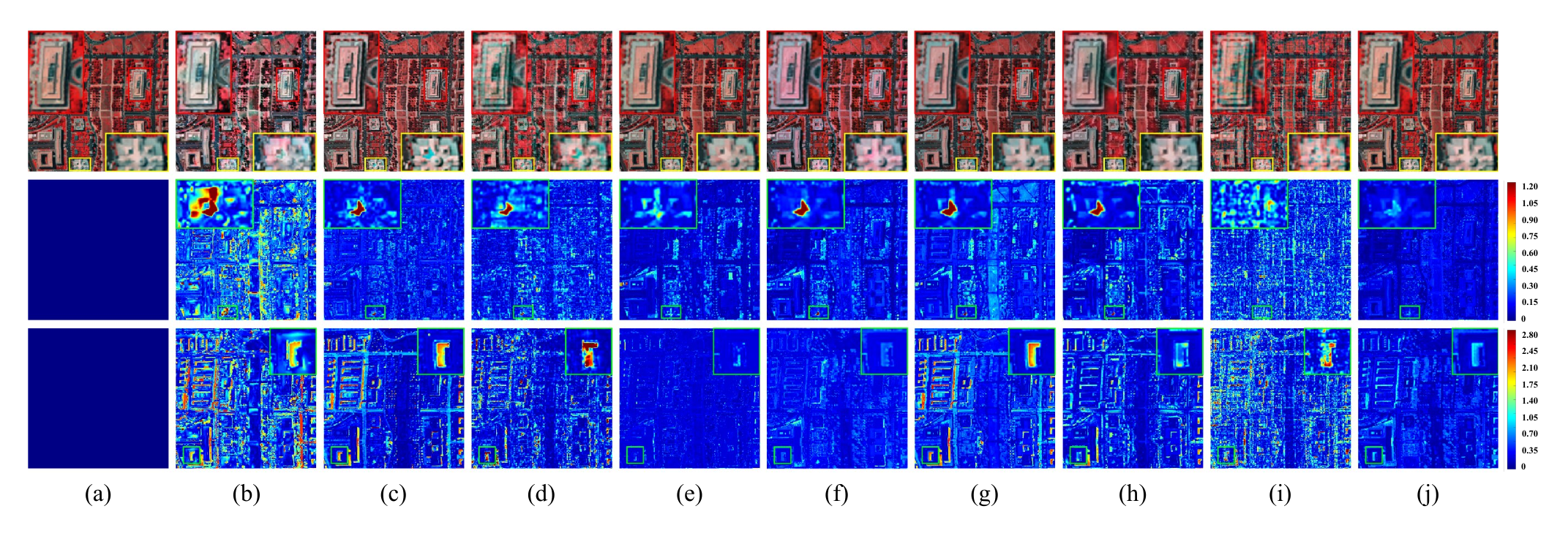}
		\caption{Illustration of fusion results on WaDC datasets. First row: pseudo-RGB image (R:60, G:27, B:17). The second and third rows are heatmaps of RMSE and SAM between the ground truth and reconstructed respectively. The error map range are [0, 1.2] and [0, 2.8]. (a).GT, (b).CNMF\cite{5982386}, (c).HySure\cite{7000523}, (d).CSTF\cite{8359412}, (e).uSDN\cite{uSDN}, (f).CuCANet\cite{10.1007/978-3-030-58526-6_13}, (g).HyCoNet\cite{HyCoNet}, (h).MIAE\cite{9681709}, (i).UDTN\cite{10115230}, (j).DTDNML}
		\label{WaDC_compare}
	\end{figure*}

\begin{figure*}
		\centering
		\setlength{\abovecaptionskip}{-0.1cm}
		\includegraphics[width=0.98\linewidth]{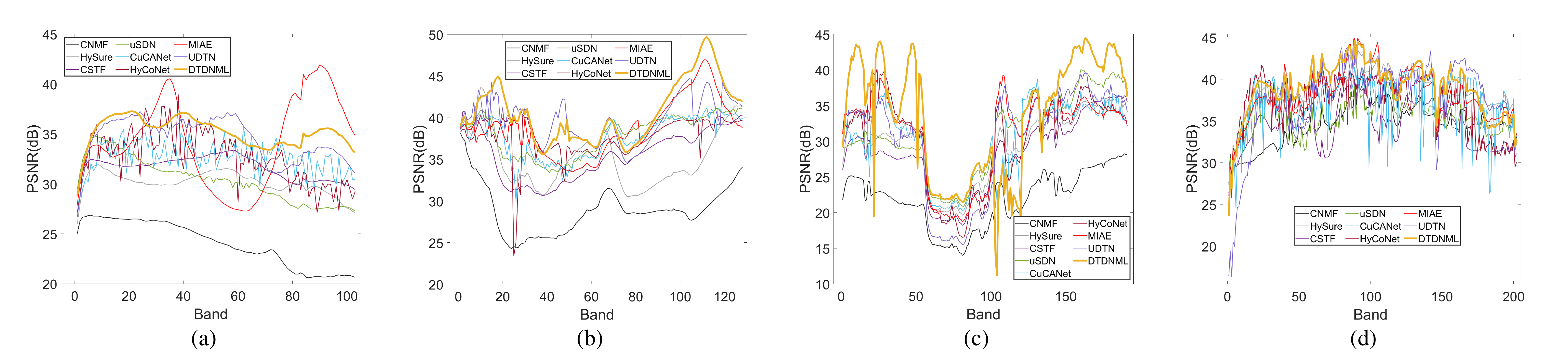}
		\caption{PSNR curves of each band in different HSIs with SR of 8. (a).Pavia University, (b).Chikusei, (c).WaDC, (d).SanDiego}
		\label{PSNR_curve}
	\end{figure*}

	\vspace{0.2cm}
	\textit{2) Chikusei}: Fig.\ref{Chikusei_compare}. illustrates the results generated by different fusion methods on the Chikusei dataset at a sampling ratio of 8. Even if each of method exhibits good performance in exploiting spatial information, there is a noticeable difference in the spectral dimension. From the results, MIAE stands out, although, in the extraction and preservation of spectral information in the lower-left corner river region among them, there is significant spectral distortion in the upper right house area. In general, however, DTDNML outperforms HyCoNet, MIAE and UDTN in overall performance. We present a detailed analysis of the metrics in the Table \ref{tab:Tab2}. It is clearly found that both HSIs achieve satisfactory results across all the metrics. 
	\begin{table}[H]
		\caption{EVALUATION INDICES OF DIFFERENT HSI–MSI FUSION METHODS ON CHIKUSEI DATASET}\label{tab:Tab2}
		\centering
		\resizebox{\linewidth}{!}{
			\begin{tabular}{ccccccc}
				\hline\hline
				\multicolumn{1}{c|}{Metrics}    & \multicolumn{1}{c|}{RMSE}            & \multicolumn{1}{c|}{PSNR}           & \multicolumn{1}{c|}{SAM}           & \multicolumn{1}{c|}{ERGAS}         & \multicolumn{1}{c|}{SSIM}           & UIQI           \\ \hline
				\multicolumn{1}{c|}{Optimal Value} & \multicolumn{1}{c|}{0}               & \multicolumn{1}{c|}{$+\infty$}      & \multicolumn{1}{c|}{0}             & \multicolumn{1}{c|}{0}             & \multicolumn{1}{c|}{1}              & 1              \\ \hline
				\multicolumn{7}{c}{SR=4}                                                                                                                                                                                                                      \\ \hline
				\multicolumn{1}{c|}{CNMF}       & \multicolumn{1}{c|}{0.0101}          & \multicolumn{1}{c|}{36.64}          & \multicolumn{1}{c|}{3.76}          & \multicolumn{1}{c|}{2.24}          & \multicolumn{1}{c|}{{\ul 0.975}}    & 0.982          \\
				\multicolumn{1}{c|}{HySure}     & \multicolumn{1}{c|}{0.0113}          & \multicolumn{1}{c|}{36.08}          & \multicolumn{1}{c|}{3.89}          & \multicolumn{1}{c|}{2.27}          & \multicolumn{1}{c|}{0.929}          & 0.984          \\
				\multicolumn{1}{c|}{CSTF}       & \multicolumn{1}{c|}{0.0102}          & \multicolumn{1}{c|}{37.49}          & \multicolumn{1}{c|}{4.78}          & \multicolumn{1}{c|}{1.99}          & \multicolumn{1}{c|}{0.948}          & 0.976          \\
				\multicolumn{1}{c|}{uSDN}       & \multicolumn{1}{c|}{0.0082}          & \multicolumn{1}{c|}{38.67}          & \multicolumn{1}{c|}{5.61}          & \multicolumn{1}{c|}{1.84}          & \multicolumn{1}{c|}{0.971}          & 0.985          \\
				\multicolumn{1}{c|}{CuCaNet}    & \multicolumn{1}{c|}{0.0072}          & \multicolumn{1}{c|}{40.96}          & \multicolumn{1}{c|}{3.49}          & \multicolumn{1}{c|}{1.68}          & \multicolumn{1}{c|}{0.974}          & 0.990          \\
				\multicolumn{1}{c|}{HyCoNet}    & \multicolumn{1}{c|}{0.0075}          & \multicolumn{1}{c|}{40.57}          & \multicolumn{1}{c|}{\textbf{3.13}} & \multicolumn{1}{c|}{1.76}          & \multicolumn{1}{c|}{0.970}          & {\ul 0.991}    \\
				\multicolumn{1}{c|}{MIAE}       & \multicolumn{1}{c|}{0.0071}          & \multicolumn{1}{c|}{40.53}          & \multicolumn{1}{c|}{{\ul 3.23}}    & \multicolumn{1}{c|}{1.98}          & \multicolumn{1}{c|}{0.968}          & 0.989          \\
				\multicolumn{1}{c|}{UDTN}       & \multicolumn{1}{c|}{{\ul 0.0066}}    & \multicolumn{1}{c|}{{\ul 41.73}}    & \multicolumn{1}{c|}{3.61}          & \multicolumn{1}{c|}{{\ul 1.56}}    & \multicolumn{1}{c|}{0.971}          & 0.985          \\
				\multicolumn{1}{c|}{DTDNML}     & \multicolumn{1}{c|}{\textbf{0.0064}} & \multicolumn{1}{c|}{\textbf{42.27}} & \multicolumn{1}{c|}{3.42}          & \multicolumn{1}{c|}{\textbf{1.52}} & \multicolumn{1}{c|}{\textbf{0.976}} & \textbf{0.992} \\ \hline
				\multicolumn{7}{c}{SR=8}                                                                                                                                                                                                                      \\ \hline
				\multicolumn{1}{c|}{CNMF}       & \multicolumn{1}{c|}{0.0264}          & \multicolumn{1}{c|}{28.92}          & \multicolumn{1}{c|}{6.71}          & \multicolumn{1}{c|}{4.87}          & \multicolumn{1}{c|}{0.910}          & 0.927          \\
				\multicolumn{1}{c|}{HySure}     & \multicolumn{1}{c|}{0.0152}          & \multicolumn{1}{c|}{35.17}          & \multicolumn{1}{c|}{4.38}          & \multicolumn{1}{c|}{2.45}          & \multicolumn{1}{c|}{{\ul 0.964}}    & 0.960          \\
				\multicolumn{1}{c|}{CSTF}       & \multicolumn{1}{c|}{0.0117}          & \multicolumn{1}{c|}{35.51}          & \multicolumn{1}{c|}{4.26}          & \multicolumn{1}{c|}{2.55}          & \multicolumn{1}{c|}{0.931}          & 0.976          \\
				\multicolumn{1}{c|}{uSDN}       & \multicolumn{1}{c|}{0.0088}          & \multicolumn{1}{c|}{37.94}          & \multicolumn{1}{c|}{6.21}          & \multicolumn{1}{c|}{2.06}          & \multicolumn{1}{c|}{0.963}          & 0.976          \\
				\multicolumn{1}{c|}{CuCaNet}    & \multicolumn{1}{c|}{0.0084}          & \multicolumn{1}{c|}{38.69}          & \multicolumn{1}{c|}{4.22}          & \multicolumn{1}{c|}{1.99}          & \multicolumn{1}{c|}{0.962}          & 0.979          \\
				\multicolumn{1}{c|}{HyCoNet}    & \multicolumn{1}{c|}{0.0094}          & \multicolumn{1}{c|}{38.19}          & \multicolumn{1}{c|}{\textbf{3.28}} & \multicolumn{1}{c|}{2.16}          & \multicolumn{1}{c|}{0.961}          & {\ul 0.988}    \\
				\multicolumn{1}{c|}{MIAE}       & \multicolumn{1}{c|}{{\ul 0.0076}}          & \multicolumn{1}{c|}{40.42}          & \multicolumn{1}{c|}{{\ul 3.31}}    & \multicolumn{1}{c|}{1.98}          & \multicolumn{1}{c|}{0.943}          & {\ul 0.988}    \\
				\multicolumn{1}{c|}{UDTN}       & \multicolumn{1}{c|}{{0.0081}}    & \multicolumn{1}{c|}{{\ul 40.65}}    & \multicolumn{1}{c|}{4.63}          & \multicolumn{1}{c|}{{\ul 1.77}}    & \multicolumn{1}{c|}{0.962}          & 0.982          \\
				\multicolumn{1}{c|}{DTDNML}     & \multicolumn{1}{c|}{\textbf{0.0075}} & \multicolumn{1}{c|}{\textbf{40.71}} & \multicolumn{1}{c|}{3.82}          & \multicolumn{1}{c|}{\textbf{1.75}} & \multicolumn{1}{c|}{\textbf{0.971}} & \textbf{0.988} \\ \hline\hline
			\end{tabular}
		}
	\end{table}
	Notably, tensor decomposition-based methods outperform matrix decomposition-based methods. The PSNR curve of each band is shown in Fig.\ref{PSNR_curve}.(b). Though curve exhibits small fluctuations between some bands, the value of PSNR is over 40 dB. For SAM, the proposed method presents shortcomings. We found that, in both cases, our method is about 0.1 to 0.4 degrees higher than the best and second best methods. One reason is that matrix decomposition-based methods extract pure endmember information through spectral unmixing, which ensures stability during the feature extraction stage. The results shows that HyCoNet performs better in spectral reconstruction than other methods, and the value of SAM can be reduced to nearly 3 degrees. On the other hand, the high-dimensional features of tensor decomposition are challenging to learn, leading to a trade-off between preserving spatial structural information and sacrificing certain spectral correlations. 
	
	\begin{figure*}
		\centering
		\setlength{\abovecaptionskip}{-0.15cm}
		\includegraphics[width=0.98\linewidth]{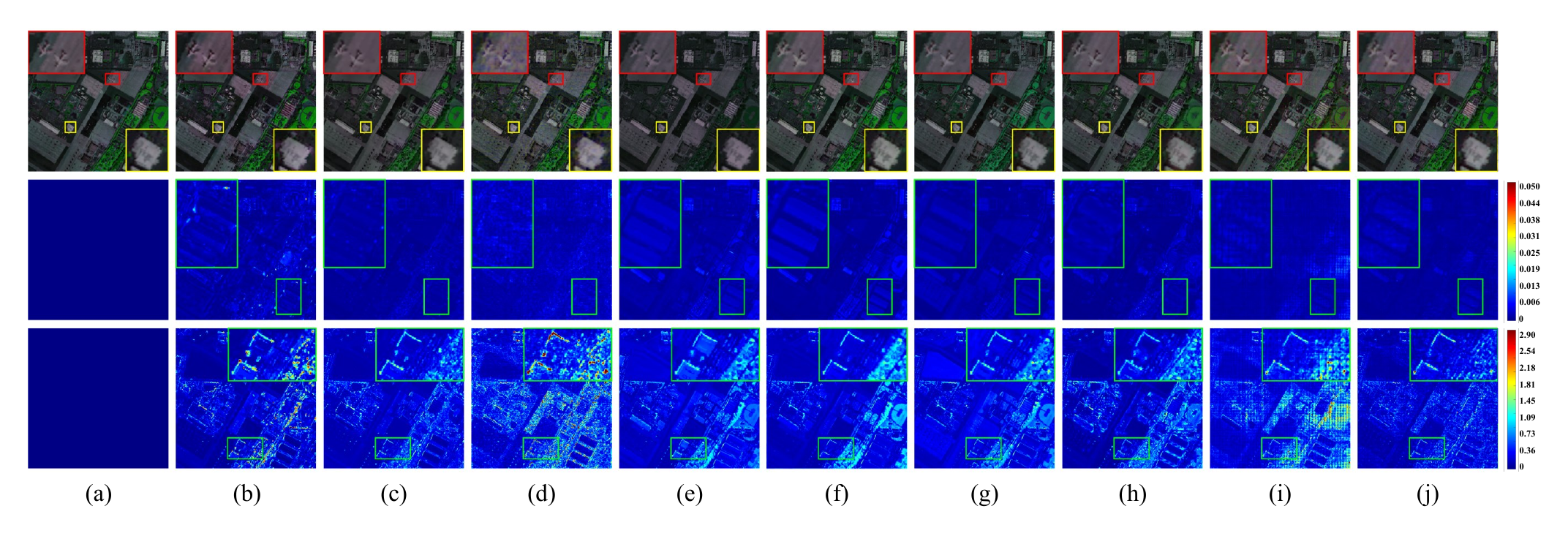}
		\caption{Illustration of fusion results on SanDiego datasets. First row: pseudo-RGB image (R:15, G:60, B:31). The second and third rows are heatmaps of RMSE and SAM between the ground truth and reconstructed respectively. The range of error maps are [0, 0.5] and [0, 2.9]. (a).GT, (b).CNMF\cite{5982386}, (c).HySure\cite{7000523}, (d).CSTF\cite{8359412}, (e).uSDN\cite{uSDN}, (f).CuCANet\cite{10.1007/978-3-030-58526-6_13}, (g).HyCoNet\cite{HyCoNet}, (h).MIAE\cite{9681709}, (i).UDTN\cite{10115230}, (j).DTDNML}
		\label{SanDiego_compare}
	\end{figure*}
	
	\vspace{0.2cm}
	\textit{3) WaDC}: The fusion results of WaDC with a sampling ratio of 8 are presented in the Fig.\ref{WaDC_compare}. It can be observed that CNMF exhibits noticeable spatial distortions, with the spatial structure of the vegetation around the White House appearing blurry. Although our proposed method achieves desirable results in the housing area compared to CuCANet and MIAE, stripe-like interferences appeared in the surrounding vegetation. This is a drawback of Conv networks for reconstruct process, which we mitigate by incorporating manifold constraints to preserve spatial structure. The results are presented in Table \ref{tab:Tab3}. 
	\begin{table}[H]
		\caption{EVALUATION INDICES OF DIFFERENT HSI–MSI FUSION METHODS ON WADC DATASET}\label{tab:Tab3}
		\centering
		\resizebox{\linewidth}{!}{
			\begin{tabular}{ccccccc}
				\hline\hline
				\multicolumn{1}{c|}{Metrics}    & \multicolumn{1}{c|}{RMSE}            & \multicolumn{1}{c|}{PSNR}           & \multicolumn{1}{c|}{SAM}           & \multicolumn{1}{c|}{ERGAS}         & \multicolumn{1}{c|}{SSIM}           & UIQI           \\ \hline
				\multicolumn{1}{c|}{Optimal Value} & \multicolumn{1}{c|}{0}               & \multicolumn{1}{c|}{$+\infty$}      & \multicolumn{1}{c|}{0}             & \multicolumn{1}{c|}{0}             & \multicolumn{1}{c|}{1}              & 1              \\ \hline
				\multicolumn{7}{c}{SR=4}                                                                                                                                                                                                                      \\ \hline
				\multicolumn{1}{c|}{CNMF}       & \multicolumn{1}{c|}{0.0330}          & \multicolumn{1}{c|}{29.67}          & \multicolumn{1}{c|}{7.45}    & \multicolumn{1}{c|}{2.16}          & \multicolumn{1}{c|}{0.929}    & 0.979          \\
				\multicolumn{1}{c|}{HySure}     & \multicolumn{1}{c|}{0.0331}          & \multicolumn{1}{c|}{30.26}          & \multicolumn{1}{c|}{7.23}          & \multicolumn{1}{c|}{2.07}          & \multicolumn{1}{c|}{0.868}          & 0.981          \\
				\multicolumn{1}{c|}{CSTF}       & \multicolumn{1}{c|}{0.0327}          & \multicolumn{1}{c|}{31.39}          & \multicolumn{1}{c|}{7.21}          & \multicolumn{1}{c|}{2.36}          & \multicolumn{1}{c|}{0.921}          & 0.982          \\
				\multicolumn{1}{c|}{uSDN}       & \multicolumn{1}{c|}{0.0347}          & \multicolumn{1}{c|}{32.03}          & \multicolumn{1}{c|}{{\ul 7.17}}          & \multicolumn{1}{c|}{{\bf 1.95}}          & \multicolumn{1}{c|}{0.930}          & 0.980          \\
				\multicolumn{1}{c|}{CuCaNet}    & \multicolumn{1}{c|}{{\ul 0.0304}}    & \multicolumn{1}{c|}{{\ul 32.59}} & \multicolumn{1}{c|}{\textbf{6.97}} & \multicolumn{1}{c|}{2.01} & \multicolumn{1}{c|}{\textbf{0.950}} & \textbf{0.985} \\
				\multicolumn{1}{c|}{HyCoNet}    & \multicolumn{1}{c|}{0.0321} & \multicolumn{1}{c|}{31.71}          & \multicolumn{1}{c|}{7.22}          & \multicolumn{1}{c|}{2.18}          & \multicolumn{1}{c|}{0.940}          & {\ul 0.984}    \\
				\multicolumn{1}{c|}{MIAE}       & \multicolumn{1}{c|}{0.0361}          & \multicolumn{1}{c|}{32.01}          & \multicolumn{1}{c|}{7.84}          & \multicolumn{1}{c|}{2.10}          & \multicolumn{1}{c|}{0.909}          & 0.983          \\
				\multicolumn{1}{c|}{UDTN}       & \multicolumn{1}{c|}{0.0389}          & \multicolumn{1}{c|}{31.10}          & \multicolumn{1}{c|}{9.69}          & \multicolumn{1}{c|}{2.15}          & \multicolumn{1}{c|}{0.905}          & 0.979          \\
				\multicolumn{1}{c|}{DTDNML}     & \multicolumn{1}{c|}{{\bf 0.0299}}          & \multicolumn{1}{c|}{\textbf{32.88}}    & \multicolumn{1}{c|}{{\ul 7.17}}          & \multicolumn{1}{c|}{{\ul 1.98}}          & \multicolumn{1}{c|}{{\ul 0.947}}    & {\bf 0.985}    \\ \hline
				\multicolumn{7}{c}{SR=8}                                                                                                                                                                                                                      \\ \hline
				\multicolumn{1}{c|}{CNMF}       & \multicolumn{1}{c|}{0.0809}          & \multicolumn{1}{c|}{21.94}          & \multicolumn{1}{c|}{14.99}         & \multicolumn{1}{c|}{5.34}          & \multicolumn{1}{c|}{0.727}          & 0.859          \\
				\multicolumn{1}{c|}{HySure}     & \multicolumn{1}{c|}{0.0389}          & \multicolumn{1}{c|}{29.64}          & \multicolumn{1}{c|}{9.17}          & \multicolumn{1}{c|}{2.33}          & \multicolumn{1}{c|}{0.921}          & 0.944          \\
				\multicolumn{1}{c|}{CSTF}       & \multicolumn{1}{c|}{0.0451}          & \multicolumn{1}{c|}{28.29}          & \multicolumn{1}{c|}{9.79}          & \multicolumn{1}{c|}{2.66}          & \multicolumn{1}{c|}{0.867}          & 0.966          \\
				\multicolumn{1}{c|}{uSDN}       & \multicolumn{1}{c|}{0.0328}          & \multicolumn{1}{c|}{{\ul 31.18}}    & \multicolumn{1}{c|}{\textbf{6.49}} & \multicolumn{1}{c|}{{\ul 1.97}}    & \multicolumn{1}{c|}{0.932}          & {\ul 0.982}    \\
				\multicolumn{1}{c|}{CuCaNet}    & \multicolumn{1}{c|}{{\ul 0.0346}}    & \multicolumn{1}{c|}{30.88}          & \multicolumn{1}{c|}{{\ul 7.31}}    & \multicolumn{1}{c|}{2.27}          & \multicolumn{1}{c|}{{\ul 0.934}}    & {\ul 0.982}    \\
				\multicolumn{1}{c|}{HyCoNet}    & \multicolumn{1}{c|}{0.0441}          & \multicolumn{1}{c|}{31.10}          & \multicolumn{1}{c|}{10.42}         & \multicolumn{1}{c|}{2.33}          & \multicolumn{1}{c|}{0.913}          & 0.976          \\
				\multicolumn{1}{c|}{MIAE}       & \multicolumn{1}{c|}{0.0403}          & \multicolumn{1}{c|}{30.98}          & \multicolumn{1}{c|}{8.38}          & \multicolumn{1}{c|}{2.34}          & \multicolumn{1}{c|}{0.888}          & 0.980          \\
				\multicolumn{1}{c|}{UDTN}       & \multicolumn{1}{c|}{0.0564}          & \multicolumn{1}{c|}{30.27}          & \multicolumn{1}{c|}{12.51}         & \multicolumn{1}{c|}{2.85}          & \multicolumn{1}{c|}{0.838}          & 0.964          \\
				\multicolumn{1}{c|}{DTDNML}     & \multicolumn{1}{c|}{\textbf{0.0318}} & \multicolumn{1}{c|}{\textbf{31.30}} & \multicolumn{1}{c|}{9.28}          & \multicolumn{1}{c|}{\textbf{1.73}} & \multicolumn{1}{c|}{\textbf{0.935}} & \textbf{0.983} \\ \hline\hline
			\end{tabular}
		}
	\end{table}
	
	\vspace{0.2cm}
	When the sampling ratio is 4, overall, deep learning methods based on tensor representation are not as effective in spectral reconstruction as methods based on matrix decomposition. But DTDNML can achieve optimal performance in PSNR, which is only higher 0.3 dB than the suboptimal. Meanwhile, when the sampling ratio is set to 8, uSDN and CuCANet show strong competitiveness, particularly in terms of the SAM metric. We attribute this phenomenon to two main factors. First of all, the dataset itself has significant noise in some specific spectral bands, leading to instability in the fusion results. Secondly, considering that WaDC consists of 191 spectral bands, it is challenging to achieve stable learning of high-dimensional features. The preservation of high-dimensional information can only be achieved by increasing the size of the core tensor, but this also adds computational complexity. Although our method is limited in certain metrics on WaDC, the PSNR of DTDNML is better than others in most bands from Fig.\ref{PSNR_curve}.(c).
	
	\vspace{0.2cm}
	\textit{4) SanDiego}: The indices of fusion results are summarized in the Table \ref{tab:Tab4}. Considering its rich spectral information, we pay more attention on the extent of spectrum loss. From the results, it can be analysed that our method is best-performing with 41.11 dB at PSNR and 4.43 degree at SAM when sampling ratio is 4. The illustration in Fig.\ref{PSNR_curve}.(d) provides the PSNR curve of each band with sampling ratio is 8. 
	
	\begin{table}[H]
		\caption{EVALUATION INDICES OF DIFFERENT HSI–MSI FUSION METHODS ON SANDIEGO DATASET}\label{tab:Tab4}
		\centering
		\resizebox{\linewidth}{!}{
			\begin{tabular}{ccccccc}
				\hline\hline
				\multicolumn{1}{c|}{Metrics}    & \multicolumn{1}{c|}{RMSE}            & \multicolumn{1}{c|}{PSNR}           & \multicolumn{1}{c|}{SAM}           & \multicolumn{1}{c|}{ERGAS}         & \multicolumn{1}{c|}{SSIM}           & UIQI           \\ \hline
				\multicolumn{1}{c|}{Optimal Value} & \multicolumn{1}{c|}{0}               & \multicolumn{1}{c|}{$+\infty$}      & \multicolumn{1}{c|}{0}             & \multicolumn{1}{c|}{0}             & \multicolumn{1}{c|}{1}              & 1              \\ \hline
				\multicolumn{7}{c}{SR=4}                                                                                                                                                                                                                      \\ \hline
				\multicolumn{1}{c|}{CNMF}       & \multicolumn{1}{c|}{0.0099}          & \multicolumn{1}{c|}{37.38}          & \multicolumn{1}{c|}{{\ul 4.51}}    & \multicolumn{1}{c|}{\textbf{2.02}} & \multicolumn{1}{c|}{{\ul 0.967}}    & 0.980          \\
				\multicolumn{1}{c|}{HySure}     & \multicolumn{1}{c|}{0.0086}          & \multicolumn{1}{c|}{38.48}          & \multicolumn{1}{c|}{4.61}          & \multicolumn{1}{c|}{2.05}          & \multicolumn{1}{c|}{0.955}          & 0.983          \\
				\multicolumn{1}{c|}{CSTF}       & \multicolumn{1}{c|}{0.0083}          & \multicolumn{1}{c|}{38.79}          & \multicolumn{1}{c|}{4.97}          & \multicolumn{1}{c|}{2.32}          & \multicolumn{1}{c|}{0.935}          & 0.984          \\
				\multicolumn{1}{c|}{uSDN}       & \multicolumn{1}{c|}{0.0088}          & \multicolumn{1}{c|}{38.70}          & \multicolumn{1}{c|}{4.92}          & \multicolumn{1}{c|}{2.19}          & \multicolumn{1}{c|}{0.937}          & 0.979          \\
				\multicolumn{1}{c|}{CuCaNet}    & \multicolumn{1}{c|}{0.0081}          & \multicolumn{1}{c|}{39.35}          & \multicolumn{1}{c|}{4.97}          & \multicolumn{1}{c|}{2.21}          & \multicolumn{1}{c|}{{\ul 0.969}}    & 0.983          \\
				\multicolumn{1}{c|}{HyCoNet}    & \multicolumn{1}{c|}{0.0085}          & \multicolumn{1}{c|}{39.73}          & \multicolumn{1}{c|}{4.69}          & \multicolumn{1}{c|}{2.31}          & \multicolumn{1}{c|}{0.959}          & 0.985          \\
				\multicolumn{1}{c|}{MIAE}       & \multicolumn{1}{c|}{{\ul 0.0084}}    & \multicolumn{1}{c|}{39.05}          & \multicolumn{1}{c|}{4.86}          & \multicolumn{1}{c|}{2.26}          & \multicolumn{1}{c|}{0.939}          & 0.978          \\
				\multicolumn{1}{c|}{UDTN}       & \multicolumn{1}{c|}{\textbf{0.0067}} & \multicolumn{1}{c|}{{\ul 40.08}}    & \multicolumn{1}{c|}{4.73}          & \multicolumn{1}{c|}{\textbf{2.02}} & \multicolumn{1}{c|}{0.960}          & {\ul 0.984}    \\
				\multicolumn{1}{c|}{DTDNML}     & \multicolumn{1}{c|}{\textbf{0.0067}} & \multicolumn{1}{c|}{\textbf{41.11}} & \multicolumn{1}{c|}{\textbf{4.43}} & \multicolumn{1}{c|}{{\ul 2.04}}    & \multicolumn{1}{c|}{\textbf{0.974}} & \textbf{0.985} \\ \hline
				\multicolumn{7}{c}{SR=8}                                                                                                                                                                                                                      \\ \hline
				\multicolumn{1}{c|}{CNMF}       & \multicolumn{1}{c|}{0.0135}          & \multicolumn{1}{c|}{34.53}          & \multicolumn{1}{c|}{5.70}          & \multicolumn{1}{c|}{2.72}          & \multicolumn{1}{c|}{0.934}          & 0.963          \\
				\multicolumn{1}{c|}{HySure}     & \multicolumn{1}{c|}{\textbf{0.0086}} & \multicolumn{1}{c|}{{\ul 38.57}}    & \multicolumn{1}{c|}{{\ul 5.24}}    & \multicolumn{1}{c|}{{\ul 2.22}}    & \multicolumn{1}{c|}{0.945}          & {\ul 0.977}    \\
				\multicolumn{1}{c|}{CSTF}       & \multicolumn{1}{c|}{0.0144}          & \multicolumn{1}{c|}{34.42}          & \multicolumn{1}{c|}{9.61}          & \multicolumn{1}{c|}{2.97}          & \multicolumn{1}{c|}{0.845}          & 0.959          \\
				\multicolumn{1}{c|}{uSDN}       & \multicolumn{1}{c|}{0.0124}          & \multicolumn{1}{c|}{35.44}          & \multicolumn{1}{c|}{6.03}          & \multicolumn{1}{c|}{2.55}          & \multicolumn{1}{c|}{0.894}          & 0.970          \\
				\multicolumn{1}{c|}{CuCaNet}    & \multicolumn{1}{c|}{0.0108}          & \multicolumn{1}{c|}{38.25}          & \multicolumn{1}{c|}{5.73}          & \multicolumn{1}{c|}{2.47}          & \multicolumn{1}{c|}{\textbf{0.954}} & 0.978          \\
				\multicolumn{1}{c|}{HyCoNet}    & \multicolumn{1}{c|}{0.0103}          & \multicolumn{1}{c|}{38.47}          & \multicolumn{1}{c|}{6.15}          & \multicolumn{1}{c|}{2.51}          & \multicolumn{1}{c|}{0.935}          & 0.976          \\
				\multicolumn{1}{c|}{MIAE}       & \multicolumn{1}{c|}{{\ul 0.0092}}    & \multicolumn{1}{c|}{38.20}          & \multicolumn{1}{c|}{5.41}          & \multicolumn{1}{c|}{2.33}          & \multicolumn{1}{c|}{0.933}          & {\ul 0.977}    \\
				\multicolumn{1}{c|}{UDTN}       & \multicolumn{1}{c|}{0.0115}          & \multicolumn{1}{c|}{38.24}          & \multicolumn{1}{c|}{8.21}          & \multicolumn{1}{c|}{5.70}          & \multicolumn{1}{c|}{0.931}          & 0.967          \\
				\multicolumn{1}{c|}{DTDNML}     & \multicolumn{1}{c|}{\textbf{0.0086}} & \multicolumn{1}{c|}{\textbf{38.94}} & \multicolumn{1}{c|}{\textbf{5.11}} & \multicolumn{1}{c|}{\textbf{2.19}} & \multicolumn{1}{c|}{{\ul 0.947}}    & \textbf{0.980} \\ \hline\hline
			\end{tabular}
		}
	\end{table}	
	\vspace{0.25cm}
	
	\begin{figure*}
		\vspace{1cm}
		\centering
		\setlength{\abovecaptionskip}{-0.2cm}
		\includegraphics[width=0.8\linewidth]{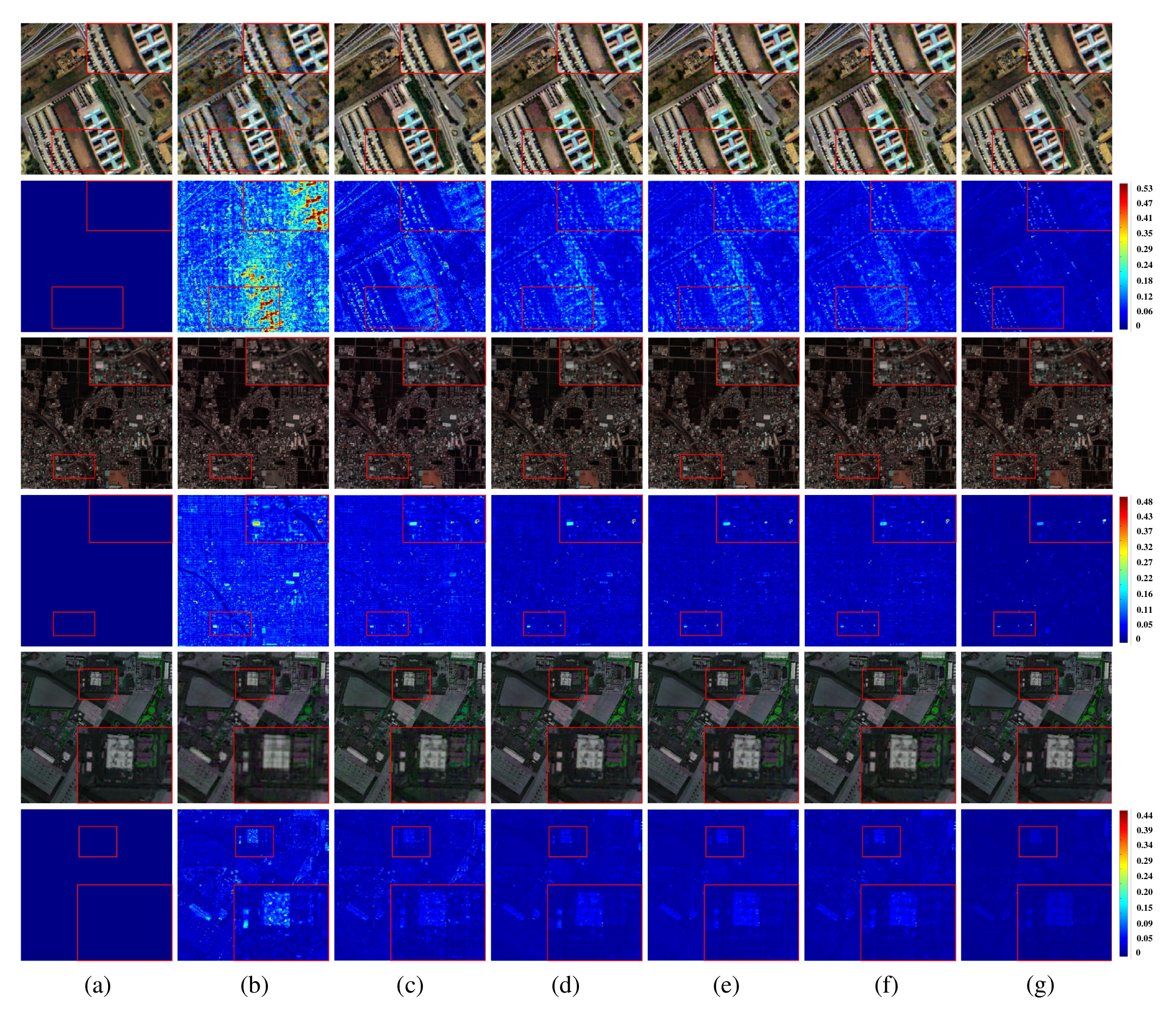}
		\caption{Ablation qualitative results on Pavia, Chikusei and SanDiego dataset. (a).GT, (b).DTDNML without CTFN, (c).DTDNML without SSAM (d).DTDNML without Spectral-Attention, (e).DTDNML without Spatial-Attention, (f).DTDNML without ML, (g).DTDNML}
		\label{ablation_qualative_compare}
	\end{figure*}
	
	Even if in some bands DTDNML get lower indices, it is stable in all bands and got the maximum average value. But tensor-based methods still have certain limitations in handling high-dimensional data, particularly in dimensionality reduction and learning of high-dimensional features. It is not difficult to find that even though our method has improved in terms of indicators, compared with matrix decomposition and spectral unmixing methods, HySure is the most challenging in terms of the loss of spatial structure and spectral information. Among the deep learning approaches, CuCANet and HyCoNet are the most challenging. To precisely observe the quality of the reconstructed images when SR is 8, the fusion results of all methods is shown in Fig.\ref{SanDiego_compare}. Each of them demonstrates satisfied fusion performance, especially in the reconstruction of larger areas such as building sections. However, some details exhibit distortion. For example, the details in the middle-right region of the image are not fully preserved, where MIAE fails to handle roof information effectively. CSTF exhibits noticeable texture loss in its results, which is also reflected in the metrics, because of the limitations on generating core tensor of this method.
	
	\vspace{-0.2cm}
	\subsection{Ablation Studies}
	In order to fully demonstrate the usefulness of our proposed CTFN with proposed methods in the experiments, we conducte ablation experiments on four datasets with two groups: the module ablation and loss function discussion. In the first group, there are five cases: \textit{1) \textbf{Case 1}}: DTDNML without CTFN, \textit{2) \textbf{Case 2}}: DTDNML without SSAM, \textit{3) \textbf{Case 3}}: DTDNML only without Spectral-Attention, \textit{4) \textbf{Case 4}}: DTDNML only without Spatial-Attention, and \textit{5) \textbf{Case 5}}: DTDNML. The first case means the use of convolutional layers to implement Tucker decomposition and obtain the core tensor instead of CTFN. The second case, on the other hand, excludes the usage of SSAM. The third and fourth cases respectively represent the exclusion of spectral or spatial attention in the proposed CTFN, primarily comparing the impact of using spectral and spatial attention mechanism alone on the fusion performance. Simultaneously, the pivotal aspect of our designed SFB, SSAB, SDAB, and SUAB lies in the interaction of attention mechanisms between modules. Hence, these three cases, to a certain extent, contribute to describing the performance of the proposed blocks. The last case is our proposed method. We give the qualitative comparison of these cases on three datasets in Fig.\ref{ablation_qualative_compare}. The indices of each case are summarized in Table \ref{tab:Tab5}.
	
	\vspace{0.2cm}
	\begin{table}[H]
		\caption{EVALUATION INDICES OF DIFFERENT MODULE ABLATION EXPERIMENTS CASES ON FOUR DATASETS}\label{tab:Tab5}
		\centering
		\resizebox{\linewidth}{!}{
			\begin{tabular}{c|cccccccc}
				\hline\hline
				\multicolumn{1}{l|}{} & \multicolumn{2}{c|}{Pavia}                                                & \multicolumn{2}{c|}{Chikusei}                                             & \multicolumn{2}{c|}{WaDC}                                                 & \multicolumn{2}{c}{SanDiego}                        \\ \cline{2-9} 
				& \multicolumn{1}{c|}{PSNR}           & \multicolumn{1}{c|}{SAM}           & \multicolumn{1}{c|}{PSNR}           & \multicolumn{1}{c|}{SAM}           & \multicolumn{1}{c|}{PSNR}           & \multicolumn{1}{c|}{SAM}           & \multicolumn{1}{c|}{PSNR}           & SAM           \\ \hline
				Case1                 & \multicolumn{1}{c|}{22.66}          & \multicolumn{1}{c|}{12.94}         & \multicolumn{1}{c|}{30.68}          & \multicolumn{1}{c|}{17.99}         & \multicolumn{1}{c|}{24.08}          & \multicolumn{1}{c|}{19.52}         & \multicolumn{1}{c|}{33.90}          & 8.12          \\
				Case2                 & \multicolumn{1}{c|}{31.89}          & \multicolumn{1}{c|}{5.90}          & \multicolumn{1}{c|}{34.66}          & \multicolumn{1}{c|}{6.12}          & \multicolumn{1}{c|}{25.93}          & \multicolumn{1}{c|}{30.72}         & \multicolumn{1}{c|}{34.26}          & 7.48          \\
				Case3                 & \multicolumn{1}{c|}{32.69}          & \multicolumn{1}{c|}{6.26}          & \multicolumn{1}{c|}{38.67}          & \multicolumn{1}{c|}{7.56}          & \multicolumn{1}{c|}{28.54}          & \multicolumn{1}{c|}{17.33}         & \multicolumn{1}{c|}{38.04}          & 5.51          \\
				Case4                 & \multicolumn{1}{c|}{32.54}          & \multicolumn{1}{c|}{6.54}          & \multicolumn{1}{c|}{39.03}          & \multicolumn{1}{c|}{5.94}          & \multicolumn{1}{c|}{28.61}          & \multicolumn{1}{c|}{14.30}         & \multicolumn{1}{c|}{38.44}          & 5.65          \\
				Case5                 & \multicolumn{1}{c|}{\textbf{34.94}} & \multicolumn{1}{c|}{\textbf{4.69}} & \multicolumn{1}{c|}{\textbf{40.71}} & \multicolumn{1}{c|}{\textbf{3.82}} & \multicolumn{1}{c|}{\textbf{31.30}} & \multicolumn{1}{c|}{\textbf{9.28}} & \multicolumn{1}{c|}{\textbf{38.94}} & \textbf{5.11} \\ \hline\hline
			\end{tabular}
		}
	\end{table}
	
	\begin{figure*}
		\centering
		\setlength{\abovecaptionskip}{-0.10cm}
		\includegraphics[width=0.9\linewidth]{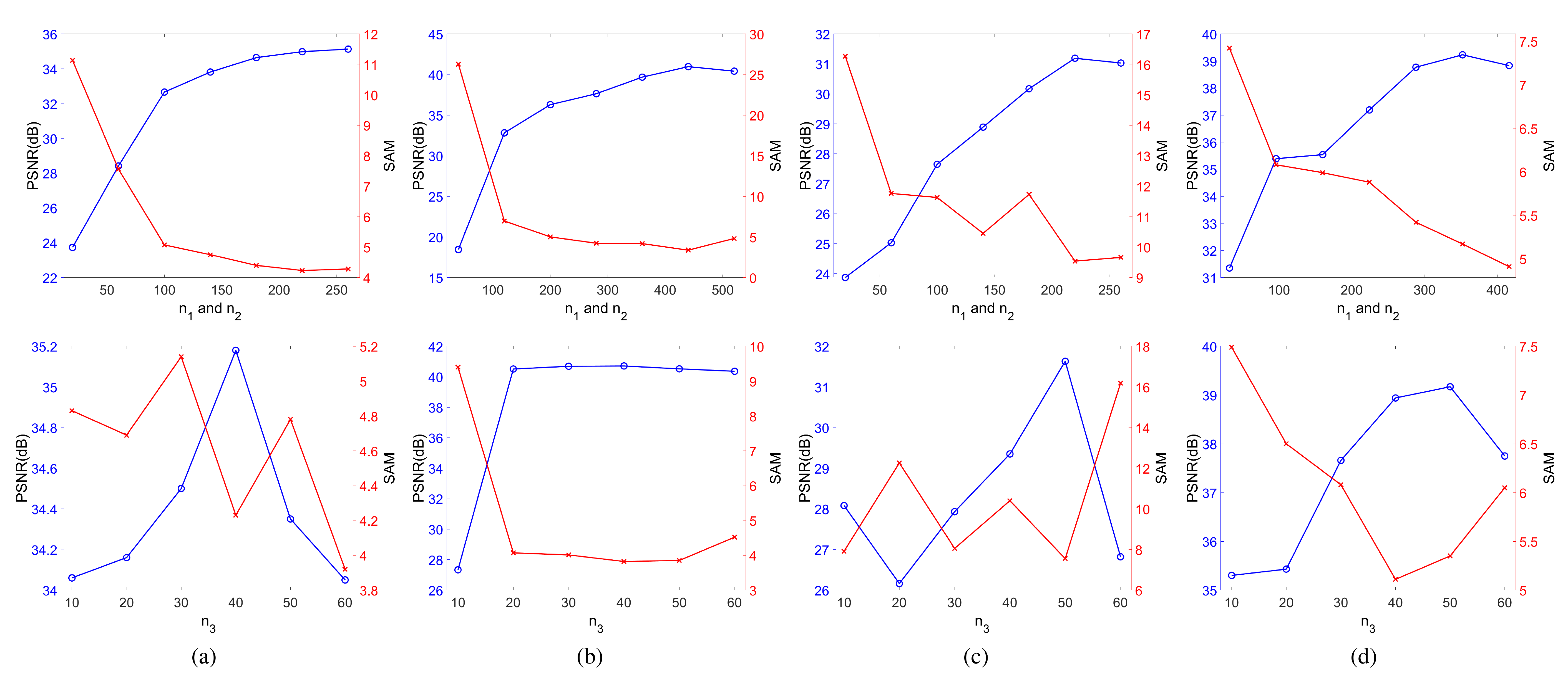}
		\caption{Indices of PSNR and SAM for the factor matrices parameters discussion in different HSIs at sampling ratio of 8. (a).Pavia University, (b).Chikusei, (c).WaDC, (d).SanDiego}
		\label{PSNR_param_wh}
	\end{figure*}
	
	\begin{figure*}
		\vspace{0.7cm}
		\centering
		\setlength{\abovecaptionskip}{-0.10cm}
		\includegraphics[width=0.95\linewidth]{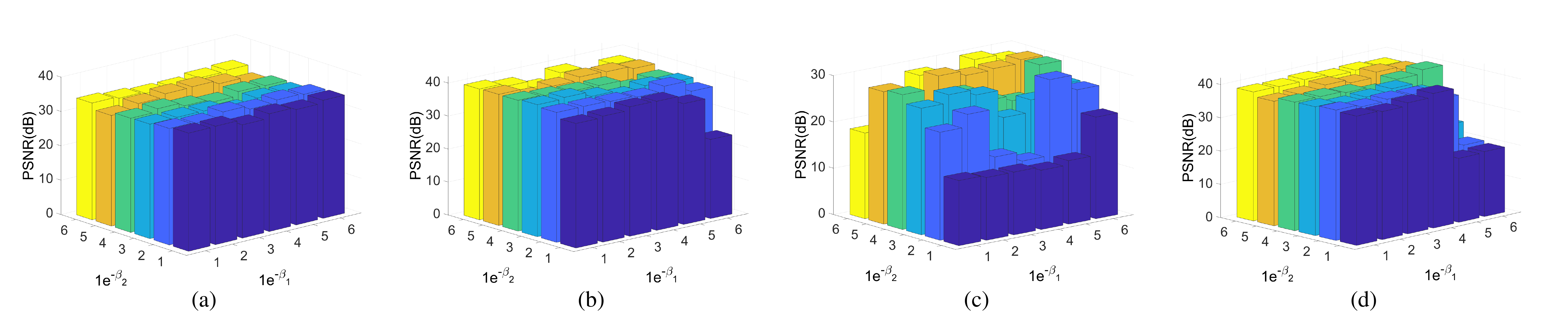}
		\caption{PSNR bars for parameters discussion on spectral and spatial manifold constraints at sampling ratio of 8. (a).Pavia University, (b).Chikusei, (c).WaDC, (d).SanDiego}
		\label{PSNR_param_beta}
	\end{figure*}
	
	The ablation experiments conducted on the four datasets revealed significant impacts of each module on the fusion super-resolution task. In the SanDiego dataset, the use of SSAM significantly improved the PSNR index by 4dB. Moreover, it also helps to reduce spectral loss. It can be clearly observed on the Pavia and Chikusei datasets that CTFN effectively improves the performance of spectral reconstruction. There is a decrease of over 7 degrees and 13 degrees in SAM, respectively. However, in the WaDC, there is a relative increase in the SAM on case 2. The reason for this phenomenon may be that the relationship between scales is not fully considered in the process of multi-scale feature learning, resulting in information loss. Comparing the experimental results of the case 3 and 4, we found that although using a single attention mechanism can effectively enhance performance, significant spectral losses still exist on the WaDC dataset. Moreover, the PSNR indicator on three datasets is approximately 1 to 2 dB lower compared to the case 5. This phenomenon emphasizes the importance of our proposed SSAM in the process of multi-scale feature fusion. 
	
	In the second group, we discussed the impact of constraint terms in the loss function on the results. It can be observed from Table.\ref{tab:tabLossItem} that the importance of reconstruction loss and degradation loss in unsupervised blind fusion tasks. $\mathcal{L}_{\text{basic}}$ is mainly used for leaning parameters of autoencoder-like network. Without it, the self-expression capability of DTDNML cannot be achieved. Simultaneously, $\mathcal{L}_{\text{PSF-SRF}}$ helps DTDNML to learn the degradation parameters and prevent overfitting. The last column of Fig.\ref{ablation_qualative_compare} demonstrate the efficient preservation of detail structure. Furthermore, we conducted ablation experiments on the proposed spectral manifold constraint. Additionally, we conducted discussions on loss training involving both L1 and L2 losses. Experimental results from Table.\ref{tab:tabLoss} indicate that, in this context, the effectiveness of L1 loss is superior.
	
	\begin{table}[H]
		\vspace{-0.2cm}
		\caption{PSNR OF DIFFERENT LOSS CONSTRAINTS ABLATION EXPERIMENTS ON FOUR DATASETS}\label{tab:tabLossItem}
		\centering
		\resizebox{\linewidth}{!}{
			\begin{tabular}{ccc|cccc}
				\hline\hline
				$\mathcal{L}_{\text{basic}}$ & $\mathcal{L}_{\text{PSF-SRF}}$ & $\mathcal{L}_{\text{Manifold}}$ & \multicolumn{1}{c|}{Pavia}  & \multicolumn{1}{c|}{Chikusei}       & \multicolumn{1}{c|}{WaDC}           & SanDiego       \\ \hline
				$\checkmark$ & $\times$ & $\times$ & \multicolumn{1}{c|}{18.79}          & \multicolumn{1}{c|}{18.26}          & \multicolumn{1}{c|}{12.90}          & 18.68          \\
				$\times$ & $\checkmark$ & $\times$ & \multicolumn{1}{c|}{21.11}          & \multicolumn{1}{c|}{19.31}          & \multicolumn{1}{c|}{16.15}          & 21.04          \\
				$\checkmark$ & $\checkmark$ & $\times$ & \multicolumn{1}{c|}{32.93}          & \multicolumn{1}{c|}{39.96}          & \multicolumn{1}{c|}{30.22}          & 38.03          \\
				$\checkmark$ & $\checkmark$ & $\checkmark$ & \multicolumn{1}{c|}{\textbf{35.18}} & \multicolumn{1}{c|}{\textbf{40.71}} & \multicolumn{1}{c|}{\textbf{31.30}} & \textbf{38.94} \\ \hline\hline
			\end{tabular}
		}
	\end{table}
	
	\begin{table}[H]
		\vspace{-0.2cm}
		\caption{INDICES OF DTDNML WITH L1 AND L2 TRAINING LOSS FUNCTIONS ON FOUR DATASETS}\label{tab:tabLoss}
		\centering
		\resizebox{\linewidth}{!}{
			\begin{tabular}{c|cc|cc|cc|cc}
				\hline\hline
				\multicolumn{1}{l|}{} & \multicolumn{2}{c|}{Pavia}     & \multicolumn{2}{c|}{Chikusei}  & \multicolumn{2}{c|}{WaDC}      & \multicolumn{2}{c}{SanDiego}   \\ \cline{2-9} 
				& \multicolumn{1}{c|}{PSNR}           & \multicolumn{1}{c|}{SAM}           & \multicolumn{1}{c|}{PSNR}           & \multicolumn{1}{c|}{SAM}           & \multicolumn{1}{c|}{PSNR}           & \multicolumn{1}{c|}{SAM}           & \multicolumn{1}{c|}{PSNR}           & SAM           \\ \hline
				L1 loss          & \multicolumn{1}{c|}{\textbf{35.18}} & \multicolumn{1}{c|}{\textbf{4.23}} & \multicolumn{1}{c|}{\textbf{40.71}} & \multicolumn{1}{c|}{\textbf{3.82}} & \multicolumn{1}{c|}{\textbf{31.30}} & \multicolumn{1}{c|}{\textbf{9.28}} & \multicolumn{1}{c|}{\textbf{38.94}} & \textbf{5.11} \\ 
				L2 loss          & \multicolumn{1}{c|}{31.60}          & \multicolumn{1}{c|}{6.89}          & \multicolumn{1}{c|}{39.16}          & \multicolumn{1}{c|}{4.28 }         & \multicolumn{1}{c|}{29.39}          & \multicolumn{1}{c|}{9.61}          & \multicolumn{1}{c|}{34.56}          & 8.76          \\ \hline\hline
			\end{tabular}
		}
	\end{table}
		
	\subsection{Parameter Discussion and Illustration}
	In order to demonstrate the reliability of our experiments, we conduct discussions regarding the parameters that need to be adjusted. There are mainly two groups: the dimension of core tensor and the manifold constraint hyperparameters settings.
	\begin{figure}[H]
		\centering
		\setlength{\abovecaptionskip}{-0.25cm}
		\includegraphics[width=\linewidth]{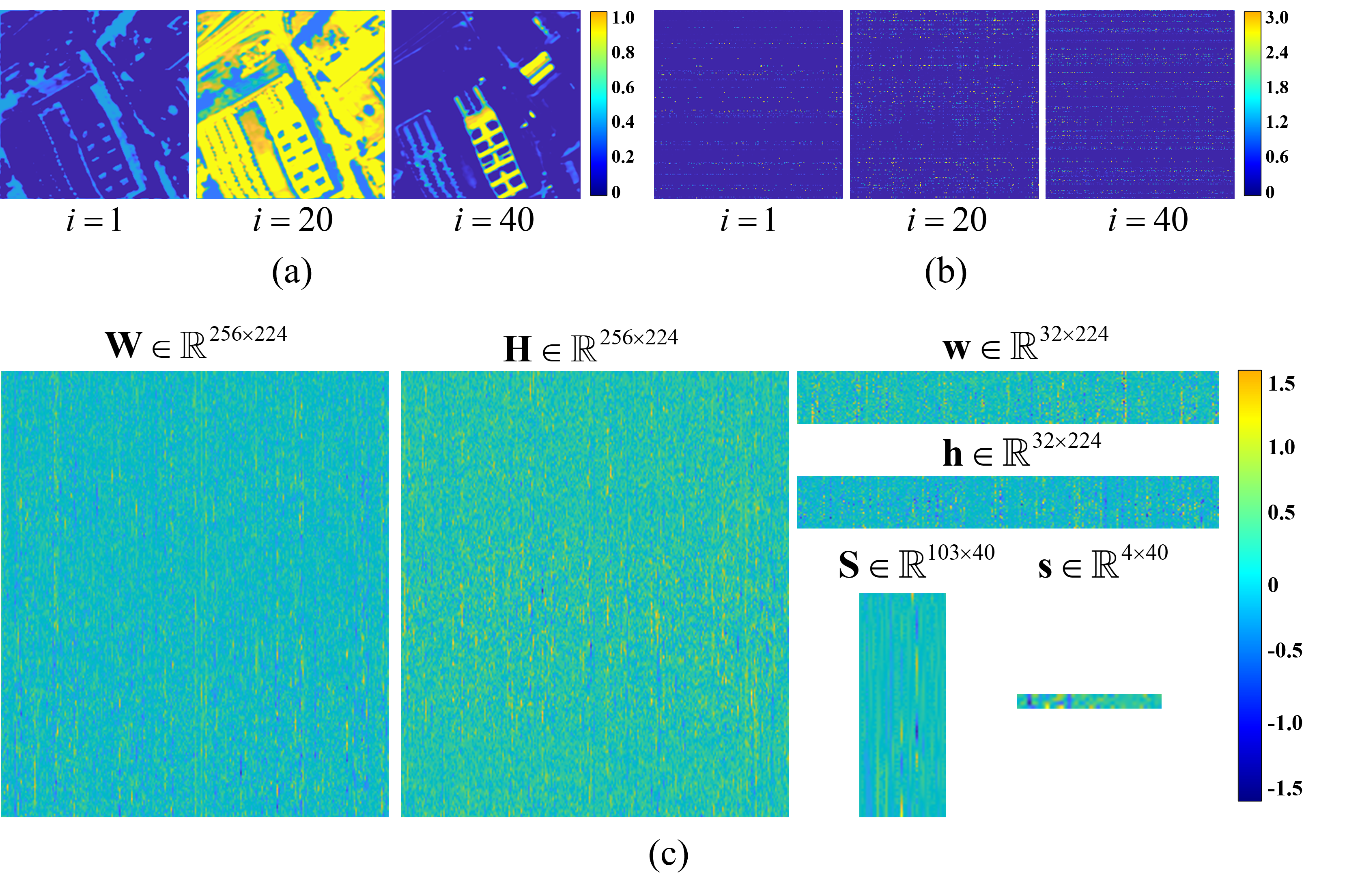}
		\caption{Illustrative results of Tucker Decomposition on the Pavia Dataset. (a).The slices of $\mathbfcal{F}_{\text{up}}$ at indices 1, 20, and 40. (b).The slices of generated core tensor at indices 1, 20, and 40. (c).The factor matrices on different modes.}
		\label{CTFM}
	\end{figure}
	For the core tensor dimension, our discussion primarily revolves around the spatial and spectral dimensions, given the consistent spatial dimensions of the simulated data used in the experiments. We organized seven groups for the the spatial dimension, and six groups for the spectral dimension. From the Fig.\ref{PSNR_param_wh}, the reduction in spectral dimensionality also effectively demonstrates the effectiveness of our low-dimensional structure learning. Alhough, the optimal dimensions of core tensor in different datasets are different, they are maintained at a certain proportion. The ratio spatial dimension reduction is around 0.85. And the spectral dimension could be reduced to 40. To effectively illustrate the output of the deep Tucker decomposition network, take the Pavia dataset as an example, we displays the slices of upsampled feature map $\mathbfcal{F}_{\text{up}}$ and core tensor, and learned factor matrices in Fig.\ref{CTFM} when the core tensor dimension is set to $224\times224\times40$. The factor matrices primarily encapsulate the base information of the three modes, while the core tensor represents a sparse order-3 coefficient matrix.
	
	Regarding the hypesrparameter discussion for the manifold constraint, we tune the hyperparameters to find the optimal settings. The values of $\beta_1$ and $\beta_2$ are set as [1e-1, 1e-2, 1e-3, 1e-4, 1e-5, 1e-6]. We illustrate the surface graph of PSNR regarding parameter selection on each dataset in Fig.\ref{PSNR_param_beta}. The performance of DTDNML is usually robust on the Pavia, Chikusei, and SanDiego datasets, with significant perturbations on WaDC. We simply set the weights to all datasets at [1e-3, 1e-2].
	
	\subsection{Computational Cost Analysis}
	
	In this section, we compare and analyze the size of parameters and inference time of proposed DTDNML with other fusion methods on CPU and GPU. For model-based methods, the CPU inference time is only provided. From the Table \ref{tab:Computation}, it is clearly observed that methods based on matrix decomposition generally have fewer parameters than tensor-based methods in most cases, but there is little difference in running time. Among them, uSDN\cite{uSDN} and HyCoNet\cite{HyCoNet} achieved the fastest. Compared to UDTN\cite{10115230}, DTDNML incurs slightly higher computational costs. However, there is negligible difference in inference time between DTDNML and other methods.
	
	\begin{table}[H]
		\vspace{-0.2cm}
		\caption{PARAMETERS(M) AND INFERENCE TIME(s) OF DIFFERENT FUSION METHODS, WHEN THE SCALING RATIO IS 8}\label{tab:Computation}
		\centering
		\resizebox{\linewidth}{!}{
			\begin{tabular}{c|cccc|cccc}
				\hline\hline
				\multirow{2}{*}{Method} & \multicolumn{4}{c|}{Parameters(M)}                                                            & \multicolumn{4}{c}{GPU/CPU Inference Time(S)}                                                                      \\ \cline{2-9} 
				& \multicolumn{1}{c|}{Pavia} & \multicolumn{1}{c|}{Chikusei}   & \multicolumn{1}{c|}{WaDC}      & SanDiego & \multicolumn{1}{c|}{Pavia}      & \multicolumn{1}{c|}{Chikusei}   & \multicolumn{1}{c|}{WaDC}       & SanDiego   \\ \hline
				CNMF                    & \multicolumn{1}{c|}{-}     & \multicolumn{1}{c|}{-}    & \multicolumn{1}{c|}{-}    & -        & \multicolumn{1}{c|}{-/6.77}     & \multicolumn{1}{c|}{-/29.51}      & \multicolumn{1}{c|}{-/7.38}       & -/17.96      \\
				HySure                  & \multicolumn{1}{c|}{-}     & \multicolumn{1}{c|}{-}    & \multicolumn{1}{c|}{-}    & -        & \multicolumn{1}{c|}{-/19.68}    & \multicolumn{1}{c|}{-/97.50}      & \multicolumn{1}{c|}{-/20.26}      & -/54.49      \\
				CSTF                    & \multicolumn{1}{c|}{-}     & \multicolumn{1}{c|}{-}    & \multicolumn{1}{c|}{-}    & -        & \multicolumn{1}{c|}{-/26.14}    & \multicolumn{1}{c|}{-/123.73}     & \multicolumn{1}{c|}{-/37.35}      & -/103.61     \\
				uSDN                    & \multicolumn{1}{c|}{0.05}  & \multicolumn{1}{c|}{0.05} & \multicolumn{1}{c|}{0.57} & 0.05     & \multicolumn{1}{c|}{0.17/1.12}  & \multicolumn{1}{c|}{0.37/2.06}  & \multicolumn{1}{c|}{0.38/1.55}  & 0.53/1.77  \\
				CuCaNet                 & \multicolumn{1}{c|}{0.45}  & \multicolumn{1}{c|}{1.24} & \multicolumn{1}{c|}{0.46} & 0.84     & \multicolumn{1}{c|}{0.18/19.73} & \multicolumn{1}{c|}{1.33/21.30} & \multicolumn{1}{c|}{0.35/24.06} & 1.47/28.71 \\
				HyCoNet                 & \multicolumn{1}{c|}{0.54}  & \multicolumn{1}{c|}{0.54} & \multicolumn{1}{c|}{0.57} & 0.57     & \multicolumn{1}{c|}{0.15/10.22} & \multicolumn{1}{c|}{1.12/15.44} & \multicolumn{1}{c|}{0.35/11.30} & 1.13/13.28 \\
				MIAE                    & \multicolumn{1}{c|}{0.06}  & \multicolumn{1}{c|}{0.06} & \multicolumn{1}{c|}{0.07} & 0.07     & \multicolumn{1}{c|}{0.22/1.74}  & \multicolumn{1}{c|}{0.24/2.15}  & \multicolumn{1}{c|}{0.31/1.76}  & 0.54/1.86  \\
				UDTN                    & \multicolumn{1}{c|}{1.06}  & \multicolumn{1}{c|}{1.52} & \multicolumn{1}{c|}{1.11} & 1.33     & \multicolumn{1}{c|}{0.20/20.21} & \multicolumn{1}{c|}{1.61/29.62} & \multicolumn{1}{c|}{0.41/21.92} & 1.32/24.81 \\
				DTDNML                  & \multicolumn{1}{c|}{1.29}  & \multicolumn{1}{c|}{1.53} & \multicolumn{1}{c|}{1.34} & 1.47     & \multicolumn{1}{c|}{0.21/22.78} & \multicolumn{1}{c|}{1.52/36.14} & \multicolumn{1}{c|}{0.41/23.90} & 1.31/26.59 \\ \hline\hline
			\end{tabular}
		}
	\end{table}	
	
	\section{Conclusion}
	In this article, we propose a HSI and MSI blind fusion method based on deep Tucker decomposition and spatial-spectral manifold learning. The core tensor is treated as a shared deep feature learned by LR-HSI and HR-MSI. We utilize theep Tucker decomposition network to generate the shared core tensor. A spatial-spectral attention mechanism is equipped in fusion module to storage multi-scale spectral and spatial information. During the reconstructed process, we transfer the mode factor matrices iterative updating into learnable decoders, and reconstruct HR-HSI by sharing parameters of networks. For the joint loss function, the blind training strategy and manifold learning is used for unsupervised learning and dimensionality reduction of deep feature representation. The effectiveness and superiority of our proposed method are validated through comparisons with existing super-resolution algorithms on publicly available datasets. However, some limitations of this work are identified and further improvements are needed through analysis of the experimental results. For instance, it is necessary to consider how to improve the compression and dimensionality reduction methods to minimize spectral loss. Additionally, it is crucial to design loss functions with better generalization performance to reduce the training cost and instability of the unsupervised model.
	
	\bibliographystyle{IEEEtran}
	\bibliography{IEEEabrv, references}
	
	\begin{IEEEbiography}[{\includegraphics[width=1in,height=1.25in,clip,keepaspectratio]{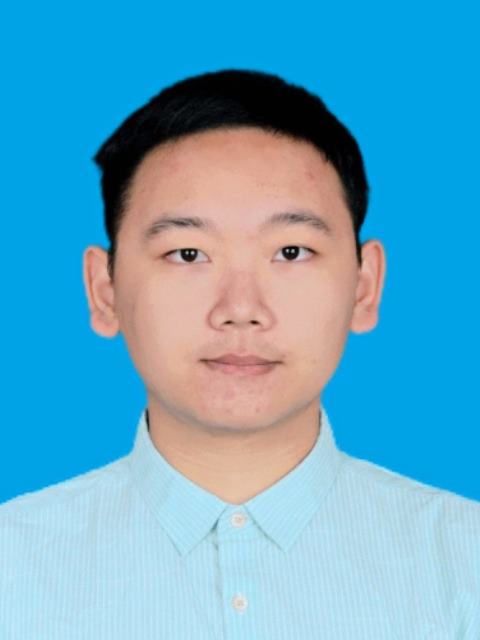}}]{He Wang} received the B.Eng. degree in data science and big data technology from Henan University (HENU), Kaifeng, Henan, China, in 2022. He is currently pursuing the Ph.D. degree with Nanjing University of Science and Technology (NJUST), Nanjing, Jiangsu, China.
		
		His research interests include hyperspectral images fusion and deep learning.
	
	\end{IEEEbiography}
	
	\vspace{-3cm}
	\begin{IEEEbiography}[{\includegraphics[width=1in,height=1.25in,clip,keepaspectratio]{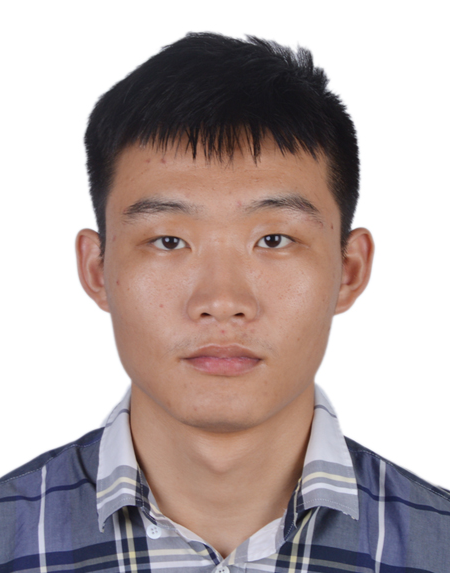}}]{Yang Xu}
		(Member, IEEE) received the B.Sc. degree in applied mathematics and the Ph.D. degree in pattern recognition and intelligence systems from Nanjing University of Science and Technology (NJUST), Nanjing, China, in 2011 and 2016, respectively.
		
		He is currently an Associate Professor with the School of Computer Science and Engineering, NUST. His research interests include hyperspectral image classification, hyperspectral detection, image processing, and machine learning.
	\end{IEEEbiography}
	
	\vspace{-3cm}
	\begin{IEEEbiography}[{\includegraphics[width=1in,height=1.25in,clip,keepaspectratio]{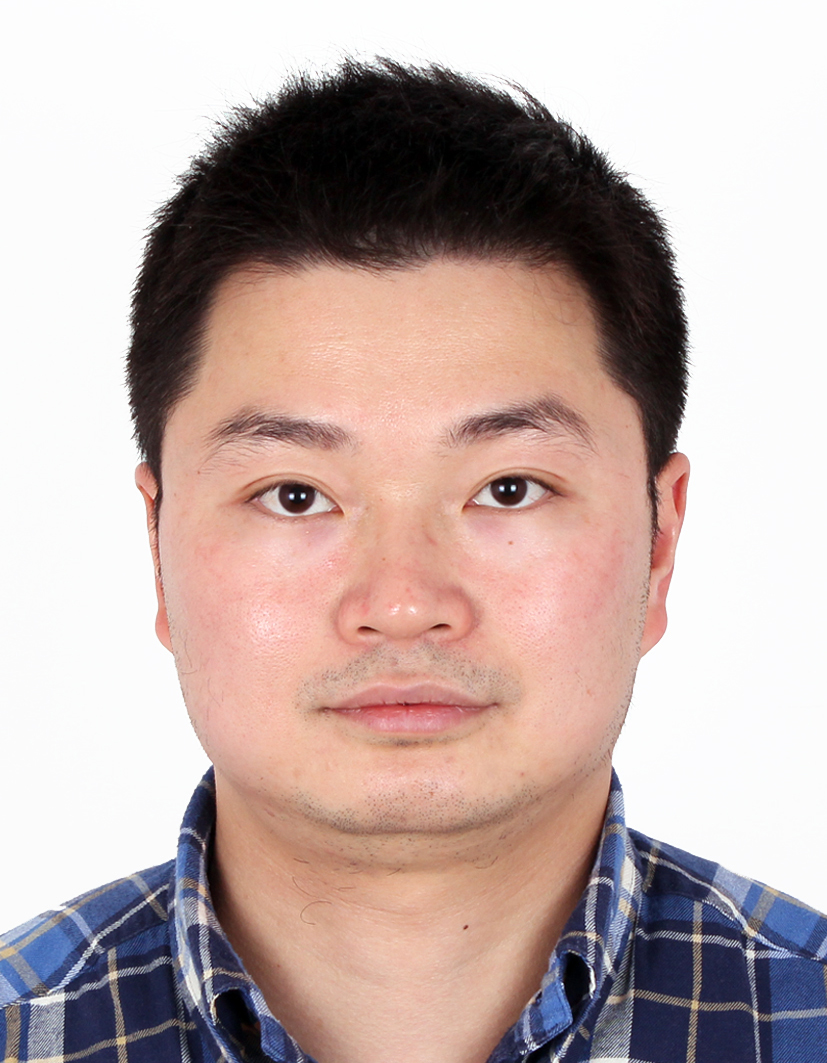}}]{Zebin Wu}
		(Senior Member, IEEE) received the B.Sc. and Ph.D. degrees in computer science and technology from Nanjing University of Science and Technology (NJUST), Nanjing, China, in 2003 and 2007, respectively. 
		
		He is currently a Professor with the School of Computer Science and Engineering, Nanjing University of Science and Technology. From August 2018 to September 2018, he was a Visiting Scholar with the GIPSA-Lab, Grenoble INP, the Université Grenoble Alpes, Grenoble, France. He was a Visiting Scholar with the Department of Mathematics, University of California at Los Angeles, Los Angeles, CA, USA, from August 2016 to September 2016 and from July 2017 to August 2017. From June 2014 to June 2015, he was a Visiting Scholar with the Hyperspectral Computing Laboratory, Department of Technology of Computers and Communications, Escuela Politécnica, University of Extremadura, Cáceres, Spain. His research interests include hyperspectral image processing, parallel computing, and big data processing.
	\end{IEEEbiography}
	
	\vspace{-3cm}
	\begin{IEEEbiography}[{\includegraphics[width=1in,height=1.25in,clip,keepaspectratio]{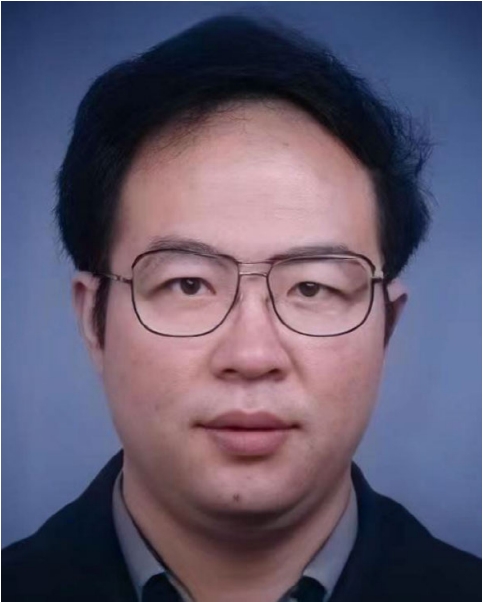}}]{Zhihui Wei}
		(Member, IEEE) was born in Jiangsu, China, in 1963. He received the B.Sc. and M.Sc. degrees in applied mathematics and the Ph.D. degree in communication and information system from Southeast University, Nanjing, China, in 1983, 1986, and 2003, respectively.
		
		He is currently a Professor and a Doctoral Supervisor with Nanjing University of Science and Technology (NJUST), Nanjing. His research interests include partial differential equations, mathematical image processing, multiscale analysis, sparse representation, and compressive sensing.
	\end{IEEEbiography}
	
\end{document}